\documentclass{article}
\usepackage[preprint]{neurips_2026}
\usepackage{graphicx} %
\usepackage{enumitem}
\setlist[itemize]{topsep=2pt}
\usepackage{booktabs}
\usepackage{array}
\usepackage{xcolor}
\usepackage{colortbl}
\usepackage{pifont}
\usepackage{url}
\usepackage{pgfplots}
\usepackage{wrapfig}
\pgfplotsset{compat=1.18}
\usepgfplotslibrary{groupplots}
\usetikzlibrary{calc}
\usepackage{longtable}

\title{Gym-Anything for Robotics: Scaling physical simulations with digital agents}
\author{Pranjal Aggarwal \qquad Sean Welleck\\Carnegie Mellon University}
\date{October 2026}

\begin{document}

\maketitle

\begin{abstract}
Realistic, fast robotics simulations could enable large-scale training of embodied AI models that transfer to the real world.
However, despite decades of progress, available simulations cover only a narrow range of real-world tasks and physical phenomena.
The core bottleneck is that creating better simulations requires significant expert effort and therefore does not scale. 
To address this, we introduce Gym-Anything for Robotics, a framework in which coding agents create realistic robotics simulations. Our core insight is that simulation creation is a coding agent task, so improvements in coding agents translate directly into richer and more faithful simulations. We select simulations based on U.S.\ GDP data to cover economically valuable physical work, define what makes a good simulation and a good task, and develop an adversarial creation-audit loop that checks each simulation and task against these definitions. Using this pipeline, we build RoboWorld, a set of 25 simulations with 25 test and 175 training RL environments of varying difficulty. Even the best frontier models fail on every RoboWorld test task, achieving an average reward of only 0.06. Building on RoboWorld, we introduce several new ways for frontier models to improve, including in-context learning across repeated attempts, a privileged teacher that reads the simulator state and teaches a student in context, self-assisted learning from the model's own privileged experience, and code policies that generate demonstrations across varied scenes to train vision-language-action models (VLAs) such as $\pi_{0.5}$. Each of these substantially improves performance.
We release our pipeline, RL environments, and benchmark to facilitate future research.\footnote{\raggedright Code and data are available at \url{https://gym-anything.github.io/robotics}.}
\end{abstract}

\begin{figure}[h]
\centering
\includegraphics[width=\linewidth]{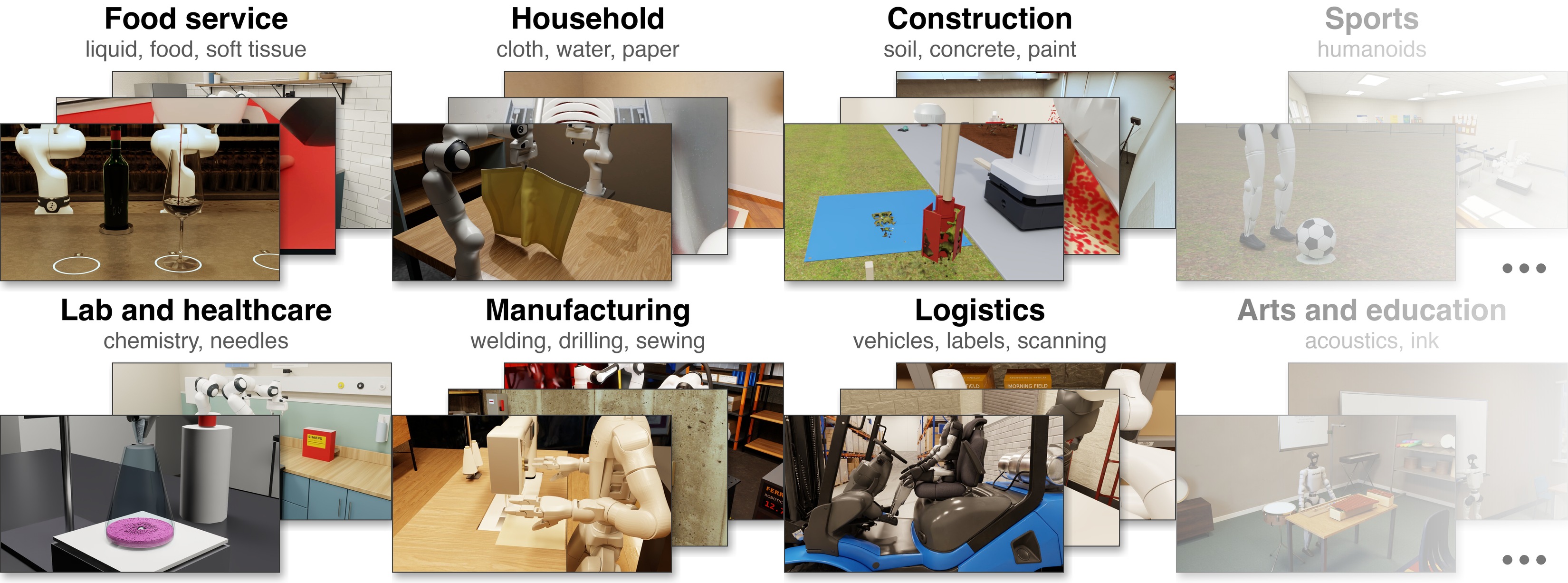}
\caption{\textbf{Built with Gym-Anything for Robotics, RoboWorld covers physical work across major sectors of the U.S.\ economy}, spanning 200 RL environments across 25 simulations.}
\label{fig:teaser}
\end{figure}

\section{Introduction}

Simulation is one of the most promising paths to training embodied AI at scale. Unlike the real world, a simulator can run thousands of episodes in parallel, reset instantly, expose the full ground-truth state, and provide rewards without human supervision~\citep{todorov2012mujoco,makoviychuk2021isaac,tao2025maniskill3}. For these advantages to matter, however, simulations must faithfully represent the real world.
Although the field has developed various hand-built environments~\citep{james2020rlbench,yu2020metaworld,liu2023libero,li2023behavior1k,nasiriany2024robocasa}, a key bottleneck is that building additional, more realistic simulations requires significant time and expert effort.
Indeed, despite decades of human progress, today's simulations cover a narrow range of real-world tasks and physical phenomena (Tables~\ref{tab:benchmarks} and~\ref{tab:prior-sims}), often with limited visual fidelity.  
Simulation as a path to embodied AI is hindered by the challenge of building diverse, realistic simulations at scale.

\begin{figure}[t]
\centering
\includegraphics[width=\linewidth]{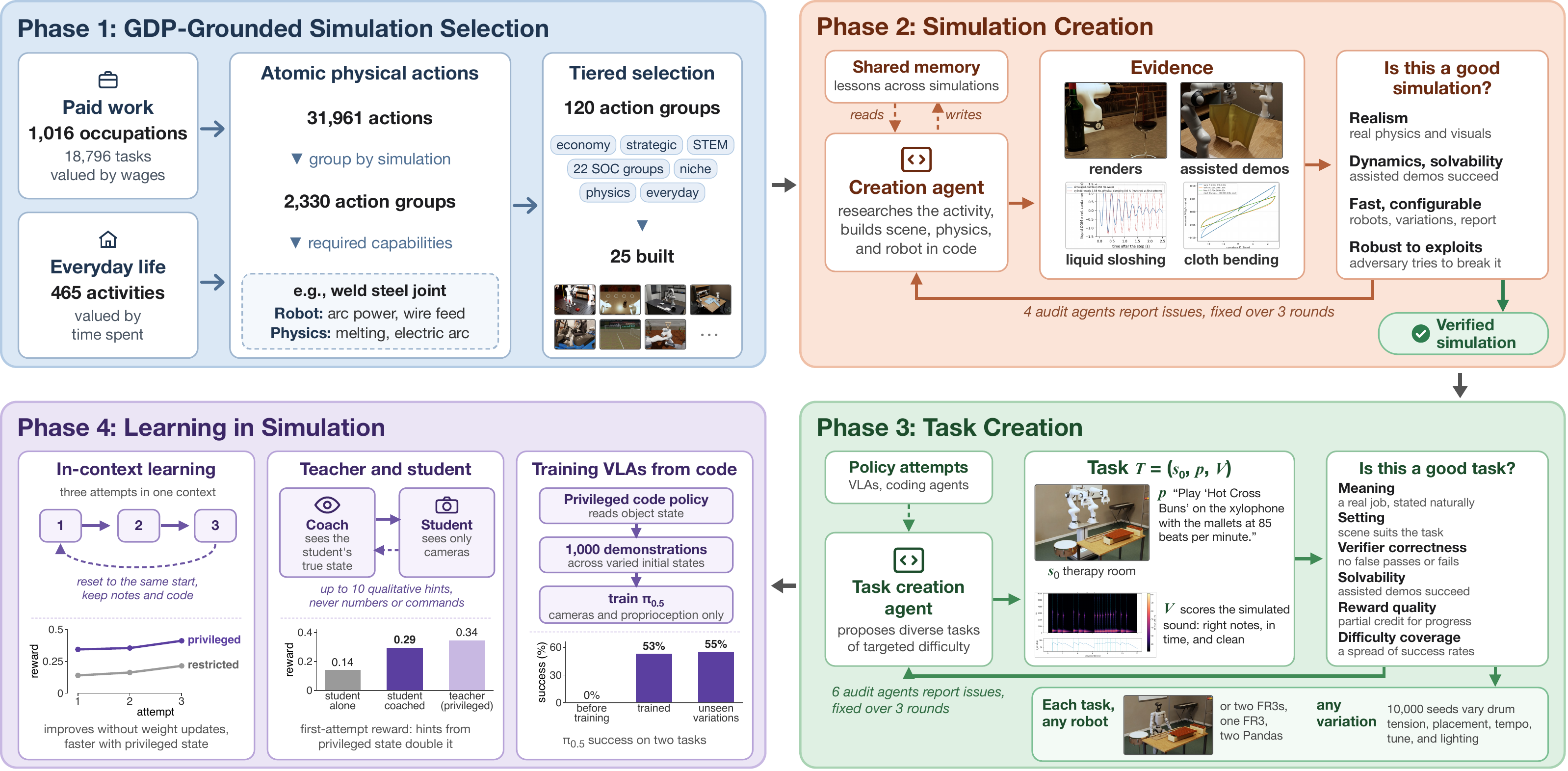}
\caption{\textbf{Overview of Gym-Anything for Robotics.} \textbf{Phase~1:} We select 120 simulations grounded in U.S.\ GDP data, covering the most valuable physical work across occupations and everyday life, and build 25 of them. \textbf{Phase~2:} A creation agent builds each simulation in code and produces evidence of its quality, which four audit agents check against our criteria. \textbf{Phase~3:} A task creation agent proposes tasks of targeted difficulty, which six audit agents check, and each task runs with many robots and variations. \textbf{Phase~4:} Models learn from privileged simulation state through in-context learning, coaching, and code policies that train VLAs.}
\label{fig:overview}
\end{figure}

Our core insight is that \textit{simulation creation can be framed as a coding agent task}. Building a simulation means writing code, including assembling assets into scenes, configuring physics, defining tasks, and implementing rewards. 
Recently, coding agents have become increasingly capable of a wide range of coding-related work~\citep{jimenez2024swebench,yang2024sweagent,wang2025openhands,wang2024gensim,wang2024robogen}.
We hypothesize that frontier agents are already capable of building diverse robotics simulations with high visual fidelity and realistic physics.
Moreover, framing simulation creation as a coding agent task allows future improvements in coding agents to translate into better simulations, thereby providing a means for scalable, realistic simulation creation that is expected to improve over time.
Building on these insights, we introduce Gym-Anything for Robotics, a framework in which coding agents create realistic robotics simulations of real-world physical work (Figure~\ref{fig:overview}).

Building simulators for real physical work requires modeling a wide variety of materials and physical phenomena.
For instance, a realistic titration requires the solution to form drops, mix, and turn pink rather than red at the endpoint. 
To organize the coverage of such phenomena in our simulation environments, we 
ground the selection of simulations in U.S.\ GDP data~\citep{aggarwal2026gymanything,gdpval2025}.
In brief, we consider each occupation, decompose the physical portion of the occupation's tasks into atomic physical actions, and weight the actions by economic value.
We then select simulations that cover the most valuable physical actions and any robotic and physics capabilities that these actions require. A coding agent then builds each simulation in code, assembles scenes from real assets, and configures the robot and physics (\S3). 

However, simulations built by coding agents frequently contain errors that make them unrealistic. For instance, a liquid simulation may not handle surface tension or inter-particle interactions correctly, or objects may pass through walls. 
Therefore, we develop an agentic \textit{creation-audit loop}~\citep{aggarwal2026gymanything} for detecting issues in the simulations and fixing them.
Specifically, we define criteria for what constitutes a good simulation, including
realism, solvability, speed, and robustness to exploits, and then run a creation agent and four audit agents in an adversarial loop.
The creation agent must produce evidence that its simulation is good, such as renders, physical measurements, and assisted demonstrations.
The four audit agents, including an adversarial agent that tries to break the simulation, then check the evidence and report issues for the creation agent to fix.

Furthermore, for each simulation, we use coding agents to create multiple tasks and control their difficulty (\S4).
This is important since tasks that are all too easy or all too hard provide little signal for evaluating or improving models~\citep{shao2024deepseekmath,yu2025dapo}. As with simulation creation, generated tasks may have issues such as being unsolvable or having incorrect rewards. We therefore define what makes a good task, and hand-curate checklists for six independent audit agents, which check each task against this definition, including solvability and reward correctness. We further ensure that each task's reward measures both progress and the final outcome, computed from the physical state of the simulation. For instance, a cleanup task can credit each object placed in its correct location rather than only a fully cleaned scene. Finally, every task can be run with different robots, from single arms to bimanual arms and humanoids, and under variations of object poses, appearance, and physics, which change the scene but not the task itself.

We use Gym-Anything for Robotics to construct RoboWorld, a collection of 25 simulations spanning domains such as manufacturing, construction, food service, laboratories, and logistics, with 25 test and 175 training RL environments of varying difficulty. RoboWorld is far from solved: even the best frontier models score 0\% on the full benchmark, while on an easier subset, the best model reaches 32\%. Beyond benchmarking, we introduce several new ways for models to learn on RoboWorld (\S5). To do so, we exploit privileged access to the full ground-truth state, which is readily available in our simulations and a benefit of the simulation paradigm. 
We first show that models can improve on RoboWorld tasks through in-context learning, and that in-context learning is accelerated by access to the privileged simulation state. Since privileged state is unavailable at test time on a real robot, we develop three ways to leverage it as a learning signal, including (1) a teacher-student setup where the teacher has privileged access and teaches a student model through qualitative hints alone, (2) a ``self-learning'' setup where a model learns from its own privileged experience, and (3) a setup in which privileged code policies train the vision-language-action model (VLA) $\pi_{0.5}$~\citep{black2025pi05}, which observes only camera images and robot proprioception.

In summary, we contribute (1) Gym-Anything for Robotics, a modular framework and multi-agent pipeline for automatically building high-quality robotics simulations, together with our definitions of what makes a good robotics simulation and a good task; (2) RoboWorld, 25 GDP-grounded simulations with 25 test and 175 training RL environments, each supporting swappable robots and variations of object poses, appearance, and physics; (3) an evaluation showing that frontier models perform poorly on RoboWorld, with no model scoring above 0\% on the full benchmark; and (4) several new ways for models to learn in simulation, showing that frontier models improve through in-context learning and do so faster with privileged state, that a frontier model with privileged state can teach a student model without it, even through verbal hints, that a model can learn from its own privileged experience, and that code policies can train VLAs such as $\pi_{0.5}$.

\section{Methodology}
\label{sec:method}

In this section, we introduce the problem setup and the GDP-grounded selection of simulations. In \S\ref{sec:creation}, we describe our multi-agent strategy for creating simulations, and in \S\ref{sec:tasks}, how we create tasks of controlled difficulty for each simulation.

\subsection{Problem Setup}
\label{sec:setup}

\begin{figure}[t]
\centering
\includegraphics[width=\linewidth]{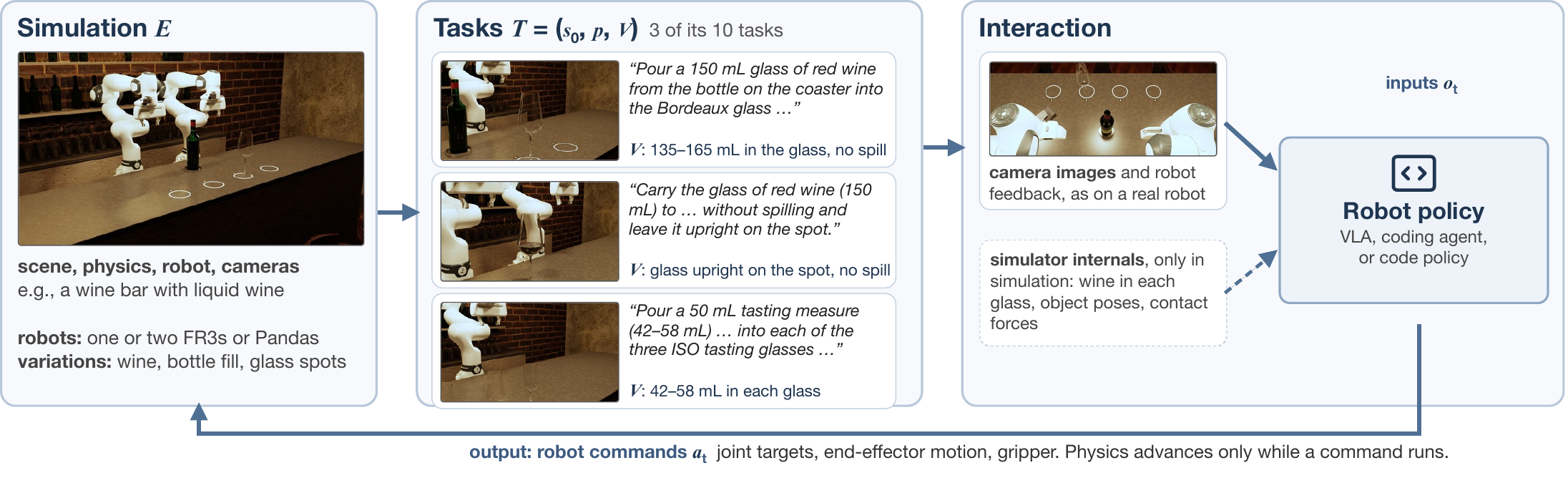}
\caption{\textbf{The problem setup, illustrated with the wine-pouring simulation.} A simulation provides the scene, physics, robots, and variations, and each of its tasks pairs an initial state with an instruction and a verifier. A robot policy observes camera images and robot feedback, and in simulation can also read internal state, such as the wine in each glass. It acts only through robot commands.}
\label{fig:setup}
\end{figure}

\paragraph{Simulation.} We refer to a simulation $E$ as a scene implemented in code, consisting of objects with physical models (such as rigid, articulated, deformable, or liquid), a robot, cameras, and a physics engine that advances the physical state $s$ of the scene (Figure~\ref{fig:setup}). Each simulation specifies a default robot, which can be swapped for other robots from a shared catalog, from single arms to bimanual arms and humanoids. Each simulation further supports variations of object poses, appearance, and physics, which change the scene in which a task is performed but not the task itself. For instance, our welding simulation is a robotic welding cell, in which a UR10e arm carrying a welding torch and an FR3 arm with a gripper work at a fixture table holding a steel plate and a bracket. Three cameras observe the cell, and the physics engine melts and solidifies the steel under the welding arc. The same cell can also be run with five other robot setups, from a single FR3 arm to a Unitree G1 humanoid, and its 10{,}000 variations shift the plate, change how far the wire sticks out of the torch, and add a draft that can blow the shielding gas away from the weld.

\paragraph{Task.} A task $T=(E_{s_0}, p, V)$ consists of a simulation $E$ with initial state $s_0$, a natural-language instruction $p$, and a verification function $V$ that maps the physical states of an episode to a binary success label and a reward $r \in [0,1]$. The reward measures both progress and outcome, and every successful completion receives $r=1$. For instance, in one welding task, $s_0$ has the bracket tacked to the plate, $p$ asks to ``weld a continuous 1/4~in fillet along the near side in the horizontal position, end to end, and fill the crater,'' and $V$ inspects the solidified weld under the AWS D1.1 welding code, checking that it is continuous and fused, reaches the specified size, and stays within the limits on undercut and porosity. If the weld fails, $r$ is proportional to the share of the required weld that is sound. Since swapping the robot or applying a variation changes only $E$ and $s_0$, this task can be run with any of the six robots and from any of the 10{,}000 variations.

\paragraph{Interaction.} At each step $t$, the agent receives an observation $o_t$, consisting of images from the robot's cameras and robot feedback such as joint positions, and issues a robot command $a_t$, such as a joint target, an end-effector motion, or a gripper command. The simulation advances physics while executing $a_t$ and returns $o_{t+1}$. Physics advances only through commands, so the time an agent spends deciding its next action does not change the scene. An episode starts from $s_0$ and ends when the agent finishes or exhausts its time budget, after which $V$ determines the score.

\paragraph{Restricted and privileged access.} Unlike the real world, a simulation can expose its full ground-truth state, such as object poses, contact forces, and material states. We therefore provide two levels of access to each simulation. Restricted access provides only what is available on a real robot: the task instruction, camera images, robot feedback, and documentation of the permitted robot commands. Privileged access additionally exposes the simulator state and the task evaluation. Unless stated otherwise, evaluated agents use restricted access, and we study how privileged access can be used for learning in \S\ref{sec:learning}.

\subsection{GDP-Grounded Simulation Selection}
\label{sec:selection}

For robots trained in simulation to be broadly useful, simulations must cover the wide variety of physical work that people value, both in their jobs and in everyday life. Similar to prior work~\citep{aggarwal2026gymanything}, we therefore ground our selection in public U.S. data (Figure~\ref{fig:overview}), combining the tasks of each occupation~\citep{peterson2001onet}, the economic value of each occupation~\citep{bls2025oews,bea2025nipa}, and the time people spend on everyday activities~\citep{bls2025atus}. However, what matters more than the tasks themselves are the atomic capabilities that they require, both from the simulation and from the robot. For instance, tightening a bolt requires a simulation that implements thread engagement and fastener friction, and a robot that can sense and limit torque. The same capabilities recur across many occupations and activities, so a simulation that provides them supports far more than a single task.

This leads to a simple question: which set of simulations provides the capabilities needed for the most valuable physical work? To answer it, we use an LLM to decompose each task into atomic physical actions, such as threading a bolt or pouring a liquid, and weight each action by the value of its task. We then group actions that a single simulation can reproduce, and identify the physics and robotic capabilities that each group requires. For instance, the action group \textit{tighten} contains 529 actions from different tasks, including tightening bolts with a wrench, driving screws, and threading spark plugs. Finally, we select action groups with an optimization that maximizes the total value of actions whose required capabilities are all supplied by the selected simulations. To ensure diversity, we run this selection in tiers covering the economic core, strategically important and STEM domains, all 22 major occupation groups, niche occupations, underrepresented capabilities such as fluids and cloth, and everyday activities. Notably, we do not restrict the selection to what current simulators support, since coding agents implement the physics each action group requires (\S\ref{sec:creation}). In total, we select 120 action groups, each defining one simulation. In this work, we build 25 of them, and we continue to add more simulations on our project website. Appendix~\ref{app:selection} describes the full pipeline.

\section{Scaling Robotics Simulations}
\label{sec:creation}

Building a realistic simulation is hard and laborious, even for experts~\citep{li2023behavior1k,nasiriany2024robocasa,wang2024robogen}. It requires researching how an activity is performed, assembling a scene from real assets, modeling the physics of the materials involved, and verifying that the result behaves physically. It often requires implementing physics that existing engines lack, such as cutting or chemical reactions. The key idea is that creating a simulation is itself a coding agent task. Building a simulation means writing code, and verifying it means running the simulation and measuring physical quantities, which coding agents can do. However, before building simulations at scale, we must define what we are building, i.e., \textit{what makes a good robotics simulation?}

To our knowledge, there is no established definition, so we define one through ten criteria in four categories. A robotics simulation must meet all ten criteria in order to be considered useful for evaluating and training robot policies:
\begin{itemize}
    \item \textbf{Realism.} (a) The simulation is physically real to the degree that is required to represent a meaningful task. For instance, a titration simulation need not model every possible chemical reaction, but the solution must form drops, mix, and turn pink rather than red at the endpoint. (b) The scene looks real: ideally, a model shown a render and a real photograph cannot tell which one is real. (c) The scene and its tasks are relevant to the real-world activities of the action group. (d) Physical parameters, such as masses and friction coefficients, are grounded in digital evidence wherever such evidence exists.
    \item \textbf{Dynamics and solvability.} (e) The interaction dynamics are realistic, such as objects settling and staying at rest, and liquids mixing. (f) Every interaction that is required by the tasks is physically achievable, as shown by assisted demonstrations in which phantom forces are allowed. For instance, for writing on a whiteboard, it suffices to show that a hand can hold the marker, the cap can be removed, the tip can leave marks under contact, and the eraser can remove them (Figure~\ref{fig:whiteboard}). Importantly, the task itself need not be solved, since our goal is to create tasks that current models cannot solve.
    \item \textbf{Fast, configurable, and documented.} (g) The simulation runs fast enough for evaluation and training, and (h) it is integrated with our library (\S\ref{sec:exp-setup}), so that its robot and scene variations can be configured. (i) For good science, a report documents its design and physical parameters with attributable, cited sources.
    \item \textbf{Robustness to exploits.} (j) No sequence of legal robot actions causes instabilities, such as non-finite values, or physically impossible states, such as objects flying away. Since reinforcement learning optimizes against the simulation, any such exploit could otherwise become learned behavior~\citep{baker2020emergent}.
\end{itemize}

\begin{figure}[t]
\centering
\includegraphics[width=\linewidth]{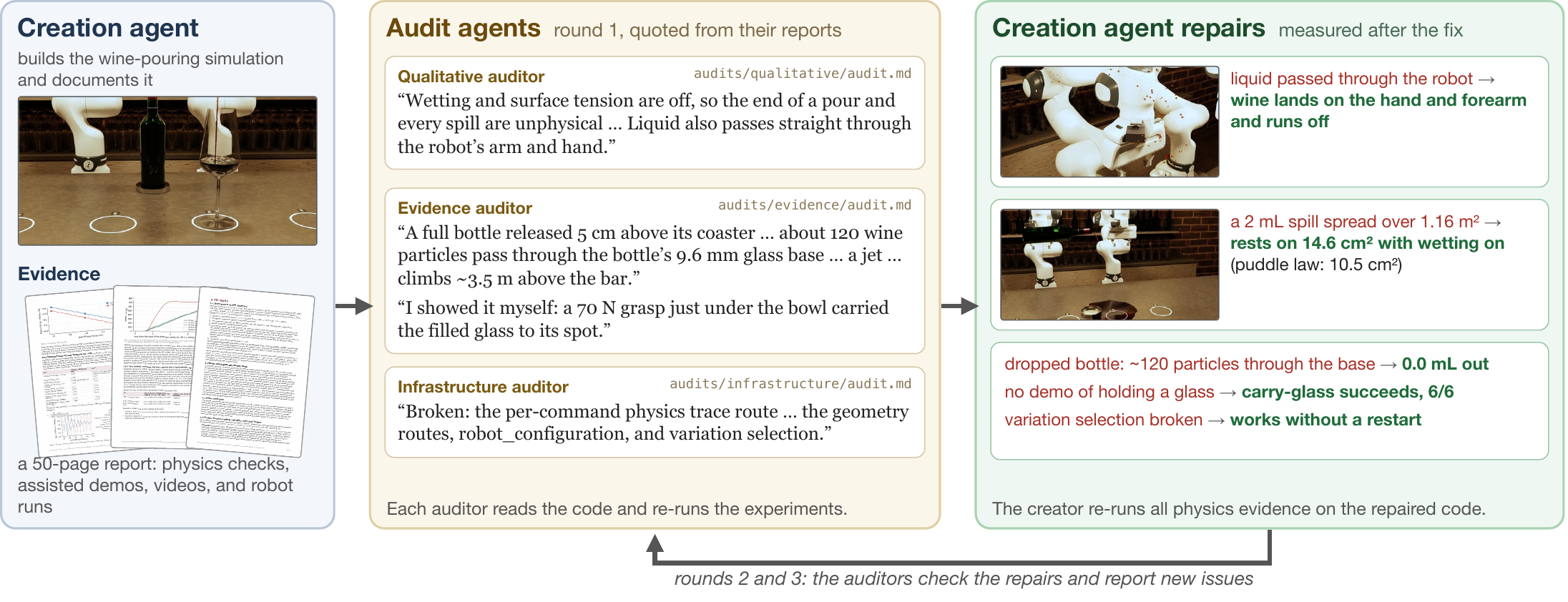}
\caption{\textbf{The creation-audit loop, shown with round 1 of the wine-pouring simulation.} The creation agent builds the simulation and a report of evidence. Audit agents read the code, re-run the experiments, and report issues, quoted here from their reports. The creation agent fixes each issue and measures the result, and the loop repeats for three rounds.}
\label{fig:creation-loop}
\end{figure}

\paragraph{Creation-audit loop.} Naively prompting even state-of-the-art coding agents results in simulations that fall short of these criteria. For instance, the agent may replace missing mechanics with animations or attachments, tune parameters until the behavior looks right, or report behavior that it never verified. Moreover, unlike software, where the application is real and a screenshot shows its true state, the physics of a simulation is written by the agent itself, so a render can show an object that has no physical effect. Similar to prior work~\citep{aggarwal2026gymanything}, we therefore use a creation-audit loop (Figure~\ref{fig:creation-loop}). A creation agent receives a selected action group, researches how its activities are performed, designs a concrete scenario, and builds the simulation together with evidence that it meets each criterion, such as videos, physical measurements, and assisted demonstrations. Four independent audit agents, one per category, then check the simulation against the criteria above, including an adversarial agent that tries to break the simulation with legal robot actions, and report issues, which the creation agent fixes. We run three such rounds for every simulation, and all agents share a memory of lessons, so that each simulation benefits from earlier ones. In the instantiation of our pipeline described here, all agents are Claude Opus 5.5~\citep{anthropic2025claude}, run via a coding agent harness such as Claude Code~\citep{anthropic2025claudecode}. Appendix~\ref{app:creation} provides further details.

\paragraph{Comparison with prior simulations.} Table~\ref{tab:prior-sims} compares three RoboWorld simulations with the closest simulations in prior benchmarks, for liquids, chemistry, and cloth. In each domain, prior works rely on a general-purpose solver and report no quantitative check of the simulated physics, while we model the specific physics that each task depends on and verify it with physics checks (Table~\ref{tab:prior-sims}). For instance, in wine pouring, the bottle glugs because of the air trapped inside it, and its emptying rate matches a measured wine bottle~\citep{kenton2012emptying}. In titration, each liquid particle carries its own chemical species, so pink appears where the titrant lands and fades as the flask is swirled. In laundry folding, garments are sewn from panels and held by the gripper only through friction. Appendix~\ref{app:comparison} extends this comparison to all 25 RoboWorld simulations.

\begin{table}[t]
\centering
\scriptsize
\newcommand{\real}[1]{\textcolor{blue!70!black}{#1}}
\caption{\textbf{RoboWorld simulations compared with the closest prior simulations.} Prior entries list what each paper reports, and -- marks nothing reported. \real{Blue} marks physics, objects, and robot limits taken from or calibrated to real data, such as measured liquid properties, manufacturer datasheets, and reference tables.}
\label{tab:prior-sims}
\setlength{\tabcolsep}{3pt}
\renewcommand{\arraystretch}{1.15}
\begin{tabular}{@{}>{\raggedright\arraybackslash}p{1.45cm}>{\raggedright\arraybackslash}p{2.9cm}>{\raggedright\arraybackslash}p{2.05cm}>{\raggedright\arraybackslash}p{2.1cm}>{\raggedright\arraybackslash}p{2.1cm}>{\raggedright\arraybackslash}p{2.25cm}@{}}
\toprule
Simulation & Physics & Objects & Robot & Success check & Physics checks \\
\midrule
\multicolumn{6}{@{}l}{\textit{Liquid: wine pouring}} \\
SoftGym\newline\citep{lin2020softgym} & position-based fluid & 1 liquid $\cdot$ box cups & none, cup moved directly & fraction poured & -- \\
ManiSkill2\newline\citep{gu2023maniskill2} & MLS-MPM & 1 liquid $\cdot$ bottle $\cdot$ beaker & 1 Panda $\cdot$ end-effector deltas & level within 4\,mm $\cdot$ under 100 particles spilt & -- \\
\textbf{Ours} & DFSPH, 2\,mm particles, 93{,}750 per bottle $\cdot$ cohesion $\cdot$ Young's-law wetting $\cdot$ lip separation $\cdot$ contact-line pinning $\cdot$ \real{headspace air and glugging} $\cdot$ added-mass coupling $\cdot$ absorbing optics & \real{4 liquids: red, white, and ros\'e wine, water} $\cdot$ \real{datasheet 75\,cl bottle, Bordeaux, ISO tasting and lined glasses, carafes} & 2 FR3 arms $\cdot$ 4 robot setups $\cdot$ \real{Franka velocity, acceleration, and jerk limits} $\cdot$ friction grasp up to 70\,N & volume in mL, e.g., 135--165\,mL $\cdot$ spill and drips in mL $\cdot$ glass upright $\cdot$ bottle returned & 12 physics checks, e.g., hydrostatic load, sloshing frequency, tipping force, pour onset $\cdot$ 12 adversarial evaluator cases \\
\midrule
\multicolumn{6}{@{}l}{\textit{Chemistry: titration}} \\
Chemistry3D\newline\citep{li2024chemistry3d} & particle fluid $\cdot$ reaction database $\cdot$ whole-solution color & 3D-scanned glassware & Franka $\cdot$ scripted pick, pour, stir & -- & -- \\
LabUtopia\newline\citep{li2025labutopia} & position-based fluid $\cdot$ LLM-inferred reactions & beakers $\cdot$ flasks & Franka $\cdot$ mobile manipulators & tilt angle past a threshold & -- \\
\textbf{Ours} & DFSPH, 348{,}160 particles $\cdot$ \real{IAPWS water} $\cdot$ surface energy $\cdot$ \real{glass and PTFE wetting} $\cdot$ 5 species per particle $\cdot$ local neutralization $\cdot$ diffusive mixing $\cdot$ \real{water equilibrium} $\cdot$ \real{indicator p$K_a$} $\cdot$ Beer--Lambert color & bored burette with PTFE stopcock $\cdot$ free flask $\cdot$ pipette with air-cushion filler $\cdot$ magnetic stir bar & 1 Panda $\cdot$ \real{manufacturer joint velocity limits} & endpoint within 1 drop $\cdot$ pink persists for 30\,s $\cdot$ spill and aliquot loss under 0.05\,mL & 8 physics checks, e.g., hydrostatic load, species and volume conservation, valve leak-tightness, force--energy consistency \\
\midrule
\multicolumn{6}{@{}l}{\textit{Cloth: laundry folding}} \\
SoftGym\newline\citep{lin2020softgym} & position-based cloth & 1 square sheet & 2 pickers & particle distance between halves & -- \\
GarmentLab\newline\citep{lu2024garmentlab} & position-based cloth & ClothesNet meshes, 11 categories & Franka, UR5, Shadow hands $\cdot$ grasp by attached blocks~\citep{wang2025dexgarmentlab} & IoU with target shape & -- \\
\textbf{Ours} & anisotropic membrane and bending $\cdot$ thickness and compression $\cdot$ static and kinetic friction $\cdot$ self-contact $\cdot$ exact intersection checks & 5 garments sewn from panels and seams: T-shirt, button-down shirt, jeans, terry towel, asymmetric top & 2 Pandas $\cdot$ finite-mass fingers $\cdot$ friction grasp, no attachments & construction-specific folds $\cdot$ garment released and at rest & 6 physics checks, e.g., free fall, fold thickness, bending release, sliding friction \\
\bottomrule
\end{tabular}
\end{table}

\paragraph{Output.} Applying this pipeline to the action groups selected in \S\ref{sec:selection}, we construct 25 simulations spanning domains such as manufacturing, construction, food service, households, laboratories and healthcare, logistics, sports, and education, with physics including liquids, chemistry, cloth, molten metal, granular soil, pastes, soft tissue, and acoustics, and robots including single arms, bimanual arms, mobile manipulators, and humanoids.

\section{Scaling Tasks}
\label{sec:tasks}

For each simulation, we create multiple tasks. Recall from \S\ref{sec:setup} that a task consists of an initial state, a natural-language instruction, and a verification function. Once a simulation is built, creating a task reduces to writing these, together with any task-specific setup of the scene. For instance, a task in the wine-pouring simulation asks: ``Pour a 150 mL glass of red wine from the bottle on the coaster into the Bordeaux glass at spot 2 (135-165 mL), without spilling or leaving drips running down the bottle, then stand the bottle back on its coaster and leave the glass upright on its spot.'' Crucially, the physics of the simulation stays fixed, since a simulation should not change depending on whether a model can complete a task. As with simulations, we first ask: \textit{what makes a good task?}

\begin{figure}[t]
\centering
\includegraphics[width=\linewidth]{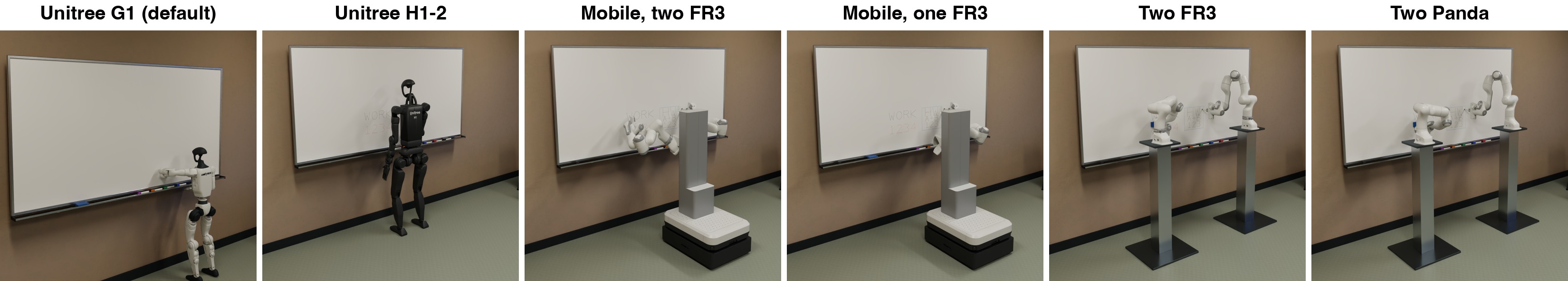}
\caption{\textbf{The whiteboard simulation with each of its six robot setups.} All six share the same classroom, markers, eraser, and physics.}
\label{fig:whiteboard}
\end{figure}

As with simulations, there is to our knowledge no established definition, so we define what a good task is through six criteria. A task must meet all six to be considered useful for evaluating and training robot policies:
\begin{itemize}
    \item \textbf{Meaning.} The task is a realistic and meaningful job, described naturally as one would describe it to a person, not as a checklist of verifier conditions.
    \item \textbf{Setting.} The scene, objects, and arrangement suit the specific task. For instance, a free kick that curls the ball requires a defensive wall, and a dribbling task requires cones.
    \item \textbf{Verifier correctness.} The verifier is deterministic and computed from the physical state of the simulation, without a VLM judge. It has neither false positives, such as accepting a cup placed over a ball for a task of putting the ball in the cup, nor false negatives, such as requiring the ball to rest at the exact center of the cup when the task did not ask for it.
    \item \textbf{Solvability.} As for simulations, assisted demonstrations show that every interaction that the task requires is physically achievable, without requiring an end-to-end solution.
    \item \textbf{Reward quality.} The reward gives continuous partial credit that reflects meaningful progress, and full credit to every valid solution, irrespective of the strategy used. For instance, a cleanup task credits each object placed in its correct location.
    \item \textbf{Difficulty coverage.} Tasks that are all too easy or all too hard provide little signal for evaluating or improving models~\citep{shao2024deepseekmath,yu2025dapo}. We therefore measure the difficulty of each task with a weaker and a stronger model, Claude Haiku 5.5 and Claude Opus 5.5, and target a spread of success rates for both. This gives a ladder of difficulty that remains informative for both weaker and stronger models.
\end{itemize}

\paragraph{Task creation-audit loop.} Similar to simulations, a task creation agent proposes tasks for each simulation, targeting five tasks per simulation that are diverse in the capabilities that they test, their difficulty, and the components of the simulation that they involve. Six independent audit agents, one per criterion, then check each task and report issues, which the task creation agent fixes, for at most three rounds. In addition, the agent uses recorded evaluation trajectories to create tasks of targeted difficulty. For instance, after observing a solver agent's (in our instantiation, Claude Haiku's) attempts at applying labels, the agent creates tasks targeting success rates of 100, 50, and 0\%: repairing a single lifted label corner, repairing several lifted regions, and picking up a printed label and applying it around a curved specimen tube, respectively. See Appendix~\ref{app:tasks} for further details.

\paragraph{An example task.} Consider the whiteboard simulation shown in Figure~\ref{fig:whiteboard}. By default, a Unitree G1 humanoid stands about a meter from a classroom whiteboard, with six capped markers and a magnetic eraser in the tray. Its hardest task gives the robot a reference drawing and asks: ``Copy the supplied colour-coded diagram of the Golden Gate Bridge onto the whiteboard, using the markers in the tray: orange towers, red cables and hangers, black roadway and blue water.'' The verifier reads only the ink left on the board and scores the copy on fidelity, cleanliness, and placement: each of the 12 elements of the diagram, such as a tower or the roadway, must be at least 90\% covered, at most 1\,cm$^2$ of ink may lie elsewhere, and the copy must sit within 30\,mm of where the reference shows it. The reward is computed from these three scores, so that, for instance, a copy with a missing cable still earns partial reward. Further, the same task can be easily varied and run with any of the six robot setups in Figure~\ref{fig:whiteboard}, such as the H1-2 humanoid or a mobile base with two FR3 arms, and under variations that move the drawing, reorder the markers, and change the robot's starting posture. Finally, the task creation agent must also show that the task is solvable. Specifically, the solvability auditor checks for demonstrations showing that every interaction the task requires is physically achievable. In these demonstrations, the robot's hand holds a marker against the board by finger contact alone, the hand holds the barrel while a phantom force pulls the cap off at 2.1\,N, and a phantom force presses each uncapped marker along all 34 strokes of the diagram. 

\begin{table}[t]
\centering
\small
\newcommand{\cmark}{\textcolor{green!50!black}{\ding{51}}}
\newcommand{\xmark}{\textcolor{red!70!black}{\ding{55}}}
\newcommand{\pcmark}[1]{\textcolor{orange!85!black}{\ding{51}$^{#1}$}}
\caption{\textbf{Comparison of RoboWorld with robotics simulation benchmarks.} \cmark~yes, \xmark~no, \textcolor{orange!85!black}{\ding{51}}~partial, following what each paper reports. Liquids and Cloth: at least one task simulates liquid or cloth, respectively. Built by AI: an AI model writes the scenes and task code. From real-world data: the tasks or simulations are chosen using surveys, activity datasets, or job data. From job data: they are chosen using occupation data, such as O*NET tasks and BLS wage data. Physics checked against real data: the paper compares simulated physical quantities with real measurements. $^{*}$An LLM proposes task outlines, and the authors write the task code. $^{\dagger}$Ego4D provides action templates, and the authors compose the tasks. $^{\ddagger}$Task types, since the paper defines no fixed set of tasks. $^{\S}$Tasks evaluated in the paper, which gives no total.}
\label{tab:benchmarks}
\setlength{\tabcolsep}{2.8pt}
\begin{tabular}{@{}lccccccc@{}}
\toprule
Benchmark & \# Tasks & Liquids & Cloth & \begin{tabular}[b]{@{}c@{}}Built\\by AI\end{tabular} & \begin{tabular}[b]{@{}c@{}}From\\real-world\\data\end{tabular} & \begin{tabular}[b]{@{}c@{}}From\\job\\data\end{tabular} & \begin{tabular}[b]{@{}c@{}}Physics\\checked against\\real data\end{tabular} \\
\midrule
\multicolumn{8}{@{}l}{\textit{Rigid objects}} \\
Meta-World~\citep{yu2020metaworld} & 50 & \xmark & \xmark & \xmark & \xmark & \xmark & \xmark \\
RLBench~\citep{james2020rlbench} & 100 & \xmark & \xmark & \xmark & \xmark & \xmark & \xmark \\
RoboTwin 2.0~\citep{chen2025robotwin} & 50 & \xmark & \xmark & \xmark & \xmark & \xmark & \xmark \\
LIBERO~\citep{liu2023libero} & 130 & \xmark & \xmark & \xmark & \pcmark{\dagger} & \xmark & \xmark \\
RoboCasa365~\citep{nasiriany2026robocasa365} & 365 & \xmark & \xmark & \pcmark{*} & \xmark & \xmark & \xmark \\
\midrule
\multicolumn{8}{@{}l}{\textit{Liquids or cloth}} \\
Chemistry3D~\citep{li2024chemistry3d} & 3$^{\ddagger}$ & \cmark & \xmark & \xmark & \xmark & \xmark & \xmark \\
ManiSkill2~\citep{gu2023maniskill2} & 20 & \cmark & \xmark & \xmark & \xmark & \xmark & \xmark \\
ARNOLD~\citep{gong2023arnold} & 8 & \cmark & \xmark & \xmark & \xmark & \xmark & \xmark \\
FluidLab~\citep{xian2023fluidlab} & 10 & \cmark & \xmark & \xmark & \xmark & \xmark & \xmark \\
LabUtopia~\citep{li2025labutopia} & 30 & \cmark & \xmark & \xmark & \xmark & \xmark & \xmark \\
RGBench~\citep{hu2025rgbench} & 3 & \xmark & \cmark & \xmark & \xmark & \xmark & \cmark \\
SoftGym~\citep{lin2020softgym} & 10 & \cmark & \cmark & \xmark & \xmark & \xmark & \xmark \\
DaXBench~\citep{chen2023daxbench} & 9 & \cmark & \cmark & \xmark & \xmark & \xmark & \xmark \\
GarmentLab~\citep{lu2024garmentlab} & 20 & \cmark & \cmark & \xmark & \xmark & \xmark & \xmark \\
LeHome~\citep{li2026lehome} & 6$^{\S}$ & \cmark & \cmark & \xmark & \xmark & \xmark & \xmark \\
BEHAVIOR-1K~\citep{li2023behavior1k} & 1{,}000 & \cmark & \cmark & \xmark & \cmark & \xmark & \xmark \\
\midrule
\multicolumn{8}{@{}l}{\textit{Built by AI}} \\
GenSim~\citep{wang2024gensim} & 100 & \xmark & \xmark & \cmark & \xmark & \xmark & \xmark \\
RoboGen~\citep{wang2024robogen} & 106 & \xmark & \xmark & \cmark & \xmark & \xmark & \xmark \\
\midrule
\textbf{RoboWorld (ours)} & \textbf{200} & \cmark & \cmark & \cmark & \cmark & \cmark & \cmark \\
\bottomrule
\end{tabular}
\end{table}

\paragraph{RoboWorld.} Applying this pipeline to the simulations constructed in \S\ref{sec:creation}, we obtain RoboWorld, a collection of 200 RL environments across 25 simulations, 25 for testing and 175 for training, each with a deterministic verifier and a continuous reward. Each of the 200 environments supports swappable robots, from single arms to bimanual arms and humanoids, and variations of object poses, appearance, and physics. These variations are not considered to be separate tasks. Rather, they give many initial states of the same task, for training policies that generalize. We further take an easier subset of 25 tasks from the training split. Table~\ref{tab:benchmarks} compares RoboWorld with existing robotics benchmarks. Several of them simulate liquids or cloth, are built by AI, or check their physics against real measurements, but RoboWorld is the only one that does all of these and also selects its simulations from job data. We release RoboWorld both as a benchmark and as RL environments for training.

\section{Learning in Simulation with Frontier Models}
\label{sec:learning}

Realistic simulations open up a new way to learn physical skills: practicing in simulation, without the cost and risk of a real robot. Prior work mostly uses simulation to train small policies with reinforcement learning, often over millions of trials per task~\citep{tobin2017domain,peng2018simtoreal,makoviychuk2021isaac}. Frontier models can learn in a different way, since they can reason about a task, write code, and improve from feedback within a single context. We therefore study how frontier models can learn in simulation, and how what they learn can be passed to the models and policies that control a real robot.

To do so, we use three features of simulation that the real world lacks: privileged access to the full ground-truth state, instant resets to the same initial state, and initial states that are easy to vary. We first study how frontier models improve on RoboWorld tasks through in-context learning, and how privileged state accelerates this improvement. Since privileged state is unavailable on a real robot, we then develop three ways to pass its benefits to models and policies that lack it: coaching through qualitative hints, learning from a model's own privileged experience, and training VLAs from privileged code policies (Figure~\ref{fig:overview}). In particular, since a simulation provides privileged state and can be run at scale, a frontier model can write code policies whose demonstrations train small policies that can be deployed on a real robot, raising the success of $\pi_{0.5}$ from 0\% to 53\% (\S\ref{sec:learning-results}).

\paragraph{In-context learning.} Since a simulation can be reset instantly, a frontier model can attempt the same task repeatedly and improve from its own failures within a single context, without any weight updates~\citep{brown2020language,shinn2023reflexion}. After each attempt, the simulation is reset to the same initial state, and the agent continues in the same conversation and workspace, retaining its observations, code, and notes from earlier attempts. We run three attempts of three hours each. To measure the value of the ground-truth state, we compare agents with restricted access against agents with privileged access, which can additionally read the simulator state during each attempt.

\paragraph{Teaching a student with privileged state.} Since privileged state is unavailable outside simulation, we next study whether a model with privileged access can teach a student model that lacks it, similar to teacher-student training with privileged information in robot learning~\citep{chen2020learning,lee2020learning}. First, a teacher model with privileged access attempts the task, without knowing that it will later teach. Once it finishes, we fork its conversation and workspace into a coach, so that the coach keeps what the teacher learned about the task. A fresh student model with restricted access then attempts the task under the same budget, and may ask the coach for at most ten hints. Before each reply, the coach reads the current privileged state of the student's simulation, so it can see mistakes the student cannot observe. It then replies with a single qualitative hint about the student's mistake or approach. Hints never contain positions, coordinates, or numerical robot commands, and the coach cannot act on the simulation, so the student must still act on its own observations. We compare coached students against solo students with the same model, budget, and initial states. In our experiments, Claude Opus 5.5 serves as both the teacher and the student.

\paragraph{Learning from privileged experience.} We further study whether a model can transfer its own privileged experience to restricted execution, without a separate teacher. We resume the conversation and workspace of a teacher that succeeded with privileged access, and ask it to perform the same task with only restricted access on new variations of the scene, such as changed initial object poses. We compare its performance against students that attempt the same tasks without privileged experience.

\paragraph{Training policies from code.} In all settings above, a frontier model controls the robot step by step, which is slow and costly. Simulation enables a different approach: a frontier model with privileged access writes a policy as code~\citep{liang2023code}, and the demonstrations of this code policy train a small VLA that can be deployed on a real robot. A code policy is a Python program that reads camera images, robot state, and privileged task-object state, and outputs only ordinary robot commands, so it cannot move objects, apply forces, change physics, or reset the simulation. For instance, a code policy removes the lid of a gift box with two arms by reading object poses and contact forces, solving the inverse kinematics of the arms, and sending joint targets. Since a code policy runs without model calls, it can be run cheaply across thousands of varied initial states. We use its successful rollouts as demonstrations of camera images and robot state, and train $\pi_{0.5}$~\citep{black2025pi05}, a VLA that observes only three camera views and robot proprioception, without any object state.

\section{Results}
\label{sec:results}

\subsection{Experimental Setup}
\label{sec:exp-setup}

\paragraph{Library.} To evaluate many kinds of policies, from frontier models to code policies and learned visuomotor policies, we build the Gym-Anything for Robotics library, which gives every simulation the same interface. As a result, any policy can run on any simulation without simulation-specific code. Behind this interface, each simulation can use its own physics engine and custom solvers, and its robot can be swapped for others, from single arms to bimanual arms and humanoids. The same simulation can also be run with restricted access, as on a real robot, or with privileged access to its full state (\S\ref{sec:setup}), so it serves evaluation, privileged experiments, and training alike. Appendix~\ref{app:library} provides further details.

\paragraph{Robot interface.} Each simulation runs as a server, and a policy connects to it over an authenticated HTTP API, whose key grants restricted or privileged access (\S\ref{sec:setup}). To support policies ranging from frontier models to code policies and VLAs, a policy can command the robot at several levels. It can set joint targets or end-effector coordinates, which a controller tracks, or send joint torques and, on humanoids, individual motor commands. Like a computer-use agent, it can also press game-style keys, such as W, A, S, and D to move the end effector, each held for a set number of control ticks.

\paragraph{Agents.} We evaluate three frontier models with extra-high reasoning effort: GPT-6 Astra, GPT-6.1 Terra, and Claude Opus 5.5~\citep{anthropic2025claude}. Each runs as a coding agent in its own harness, Claude Code~\citep{anthropic2025claudecode} for Claude Opus 5.5 and Codex~\citep{openai2025codex} for the GPT models, and controls the robot with restricted access through a small set of tools, such as \texttt{observe}, which returns camera images and robot feedback, and \texttt{act}, which executes a robot command. The learning experiments (\S\ref{sec:learning-results}) use Claude Opus 5.5 in every role.

\paragraph{Evaluation.} Each attempt runs on an independent copy of the simulation, starting from the task's initial state, with a wall-clock budget of three hours. The verifier scores each attempt after the agent finishes.
During the attempt, agents do not observe rewards. We report two metrics: (1) success rate, i.e., the fraction of attempts that the verifier labels as successful, and (2) average reward, i.e., the mean of the continuous reward $r \in [0,1]$, which captures partial progress.

\subsection{Main Results}
\label{sec:main-results}

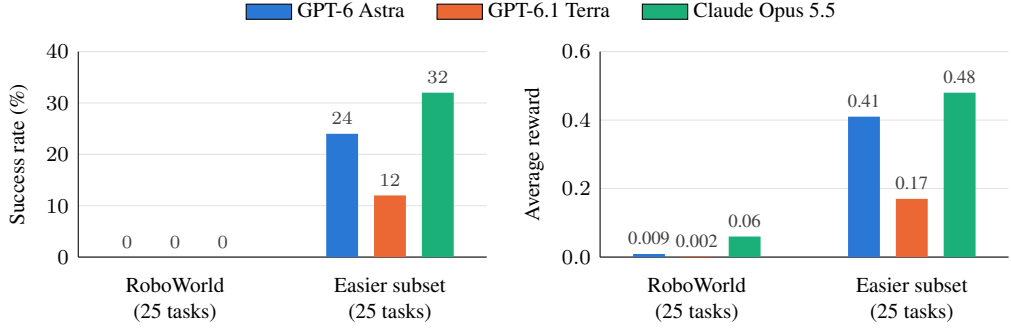
\begin{figure}[t]
\centering
\definecolor{astra}{HTML}{2A78D6}
\definecolor{terra}{HTML}{EB6834}
\definecolor{opus}{HTML}{1BAF7A}
\begin{tikzpicture}
\begin{groupplot}[
    group style={group size=2 by 1, horizontal sep=1.5cm},
    width=0.5\linewidth, height=4.3cm,
    ybar=6pt, /pgf/bar width=12pt,
    symbolic x coords={test, easy},
    xtick=data,
    xticklabels={RoboWorld\\(25 tasks), Easier subset\\(25 tasks)},
    xticklabel style={align=center},
    enlarge x limits=0.45,
    ymin=0,
    axis x line*=bottom, axis y line*=left,
    ymajorgrids, grid style={line width=0.3pt, draw=black!12},
    tick style={draw=none},
    tick label style={font=\footnotesize}, label style={font=\footnotesize},
    nodes near coords,
    every node near coord/.append style={font=\scriptsize, text=black!75},
]
\nextgroupplot[ylabel={Success rate (\%)}, ymax=40, ytick={0,10,20,30,40},
    legend to name=mainresultslegend, legend columns=3,
    legend style={draw=none, font=\footnotesize, /tikz/every even column/.append style={column sep=10pt}},
    area legend,
    nodes near coords={\pgfmathprintnumber[fixed, precision=0]{\pgfplotspointmeta}}]
\addplot[fill=astra, draw=none] coordinates {(test,0) (easy,24)};
\addplot[fill=terra, draw=none] coordinates {(test,0) (easy,12)};
\addplot[fill=opus, draw=none] coordinates {(test,0) (easy,32)};
\legend{GPT-6 Astra, GPT-6.1 Terra, Claude Opus 5.5}
\nextgroupplot[ylabel={Average reward}, ymax=0.6, ytick={0,0.2,0.4,0.6},
    yticklabel style={/pgf/number format/.cd, fixed, fixed zerofill, precision=1},
    point meta=explicit symbolic, nodes near coords={\pgfplotspointmeta}]
\addplot[fill=astra, draw=none] coordinates {(test,0.009) [0.009] (easy,0.41) [0.41]};
\addplot[fill=terra, draw=none] coordinates {(test,0.002) [0.002] (easy,0.17) [0.17]};
\addplot[fill=opus, draw=none] coordinates {(test,0.06) [0.06] (easy,0.48) [0.48]};
\end{groupplot}
\node[anchor=south] at ($(group c1r1.north west)!0.5!(group c2r1.north east)$) [yshift=4pt] {\pgfplotslegendfromname{mainresultslegend}};
\end{tikzpicture}
\caption{\textbf{Frontier models fail every RoboWorld test task, but succeed on some easier tasks.} Success rate (left) and average reward (right) on the 25 test tasks and on an easier subset of 25 tasks from the training split. All models use extra-high reasoning effort.}
\label{fig:main-results}
\end{figure}

\paragraph{Frontier models fail every RoboWorld test task.} Figure~\ref{fig:main-results} shows the performance of three frontier models on RoboWorld. None of them completes any of the 25 test tasks, so all three achieve a success rate of 0\%. Their average rewards of only 0.002 to 0.06 show that they make little progress within these tasks, and even the strongest model, Claude Opus 5.5, earns no reward at all on 16 of the 25 tasks. On an easier subset of 25 tasks from the training split, all three models complete some tasks, and Claude Opus 5.5 is again the strongest, with a success rate of 32\% and an average reward of 0.48, followed by GPT-6 Astra (24\%, 0.41) and GPT-6.1 Terra (12\%, 0.17). These results highlight that, while frontier models can complete some simpler tasks, they remain far from performing realistic physical work, establishing RoboWorld as a challenging benchmark for embodied AI. Moreover, since at least one model completes 44\% of the easier tasks, and the models earn partial reward on most of the rest, these tasks are in the sweet spot of difficulty for training frontier models with reinforcement learning.

\begin{table}[t]
\centering
\small
\setlength{\tabcolsep}{4pt}
\caption{\textbf{Success drops as the robot's degrees of freedom increase.} Success rate and average reward of GPT-6 Astra, GPT-6.1 Terra, and Claude Opus 5.5 on the 25 easier tasks, grouped by robot.}
\label{tab:robot-results}
\begin{tabular}{llcccccccc}
\toprule
 & & & & \multicolumn{3}{c}{Success (\%)} & \multicolumn{3}{c}{Avg.\ reward} \\
\cmidrule(lr){5-7}\cmidrule(lr){8-10}
Robot level & Robots & DoF & Tasks & Astra & Terra & Opus & Astra & Terra & Opus \\
\midrule
    Fixed-base arms & FR3, Panda, UR10e & 8 to 16 & 17 & 29 & 18 & 35 & 0.47 & 0.19 & 0.55 \\
    Mobile manipulator & Mobile FR3 Duo & 20 & 5 & 20 & 0 & 20 & 0.47 & 0.21 & 0.34 \\
    Humanoid & G1, H1-2 & 39 to 43 & 3 & 0 & 0 & 33 & 0.00 & 0.00 & 0.33 \\
\bottomrule
\end{tabular}
\end{table}

\paragraph{Success drops as the robot's degrees of freedom increase.} We group the 25 easier tasks into three levels by the degrees of freedom (DoF) of their robot. Fixed-base arms (17 tasks) have 8 to 16 DoF, and are mostly two Franka FR3 arms with parallel grippers. The mobile manipulator (5 tasks) is the Franka Mobile FR3 Duo, which adds a moving base and a lift to two FR3 arms, for 20 DoF. The humanoids (3 tasks) are the Unitree G1 with Dex3 hands and the Unitree H1-2 with Inspire hands, with 43 and 39 DoF. Table~\ref{tab:robot-results} shows the results at each level. With fixed-base arms, the models succeed on 18 to 35\% of tasks. With the mobile manipulator, each model succeeds on at most 20\% of tasks. With humanoids, GPT-6 Astra and GPT-6.1 Terra earn no reward at all, and Claude Opus 5.5 succeeds only in football, by kicking the ball while the robot falls (\S\ref{sec:learning-results}). These results show that controlling robots with many degrees of freedom remains a major bottleneck.

\subsection{Learning in Simulation with Frontier Models}
\label{sec:learning-results}

\begin{figure}[t]
\centering
\definecolor{solo}{HTML}{2A78D6}
\definecolor{teacher}{HTML}{EB6834}
\definecolor{coached}{HTML}{1BAF7A}
\definecolor{selftransfer}{HTML}{EDA100}
\definecolor{vla}{HTML}{6250D6}
\pgfplotsset{learningstyle/.style={
    axis x line*=bottom, axis y line*=left,
    ymajorgrids, grid style={line width=0.3pt, draw=black!12},
    tick style={draw=none},
    tick label style={font=\footnotesize}, label style={font=\footnotesize},
}}
\begin{minipage}[b]{0.46\linewidth}
\centering
\begin{tikzpicture}
\begin{axis}[learningstyle,
    width=0.78\linewidth, height=4.6cm,
    xmin=0.8, xmax=3.2, xtick={1,2,3}, xticklabels={1, 2, 3}, xlabel={Attempt},
    ymin=0, ymax=0.5, ytick={0,0.1,0.2,0.3,0.4,0.5},
    yticklabel style={/pgf/number format/.cd, fixed, fixed zerofill, precision=1},
    ylabel={Average reward},
    clip=false,
    every axis plot/.append style={line width=1.2pt, mark size=2pt},
]
\addplot[color=teacher, mark=square*] coordinates {(1,0.344) (2,0.355) (3,0.412)};
\addplot[color=coached, mark=triangle*] coordinates {(1,0.293) (2,0.309) (3,0.313)};
\addplot[color=selftransfer, mark=diamond*] coordinates {(1,0.167) (2,0.257) (3,0.287)};
\addplot[color=solo, mark=*] coordinates {(1,0.140) (2,0.163) (3,0.215)};
\node[anchor=west, font=\scriptsize, text=black!80] at (axis cs:3.1,0.412) {Teacher, 0.41};
\node[anchor=west, font=\scriptsize, text=black!80] at (axis cs:3.1,0.324) {Coached, 0.31};
\node[anchor=west, font=\scriptsize, text=black!80] at (axis cs:3.1,0.278) {Self-transfer, 0.29};
\node[anchor=west, font=\scriptsize, text=black!80] at (axis cs:3.1,0.215) {Solo, 0.22};
\end{axis}
\end{tikzpicture}
\end{minipage}%
\begin{minipage}[b]{0.54\linewidth}
\centering
\begin{tikzpicture}
\begin{axis}[learningstyle,
    width=\linewidth, height=4.6cm,
    ybar, /pgf/bar width=15pt,
    symbolic x coords={code, opus, base, trained, unseen},
    xtick={code, opus, base, trained, unseen},
    xticklabels={Code policy\\(privileged), Opus 5.5, $\pi_{0.5}$\\(base), $\pi_{0.5}$\\(trained), {$\pi_{0.5}$, unseen\\variations}},
    xticklabel style={align=center, font=\scriptsize},
    enlarge x limits=0.1,
    ymin=0, ymax=110, ytick={0,20,40,60,80,100},
    ylabel={Success rate (\%)},
    nodes near coords={\pgfmathprintnumber[fixed, precision=1]{\pgfplotspointmeta}},
    every node near coord/.append style={font=\scriptsize, text=black!75},
]
\addplot[fill=teacher, draw=none, bar shift=0pt] coordinates {(code,96.7)};
\addplot[fill=coached, draw=none, bar shift=0pt] coordinates {(opus,60)};
\addplot[fill=black!25, draw=none, bar shift=0pt] coordinates {(base,0)};
\addplot[fill=vla, draw=none, bar shift=0pt] coordinates {(trained,53) (unseen,55)};
\end{axis}
\end{tikzpicture}
\end{minipage}
\caption{\textbf{Frontier models learn in simulation, and their code policies train a small VLA.} Left: average reward of Claude Opus 5.5 over 11 RoboWorld simulations after each of three attempts. Only the teacher has privileged access during its attempts, and self-transfer resumes the teacher with restricted access. Right: success rate on two gift box tasks. A frontier model writes code policies with privileged access, and their demonstrations train $\pi_{0.5}$, which observes only camera images and robot proprioception, as on a real robot. Opus 5.5 controls the robot step by step.}
\label{fig:learning-results}
\end{figure}
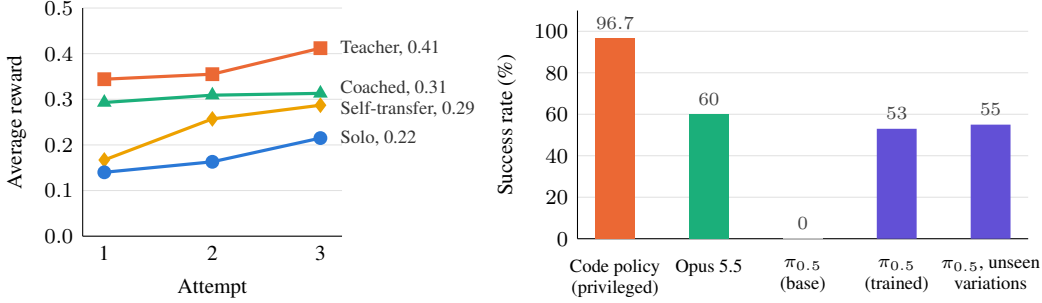

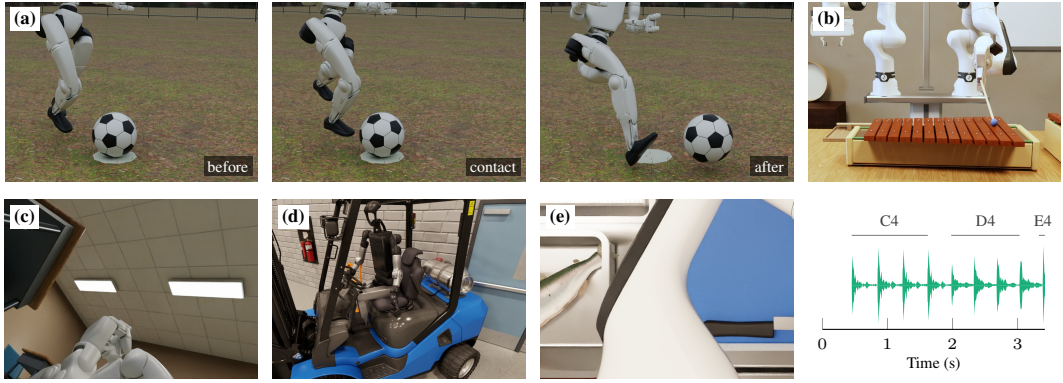
\begin{figure}[t]
\centering
\definecolor{opus}{HTML}{1BAF7A}
\newlength{\bw}\setlength{\bw}{\dimexpr(\linewidth-6mm)/4\relax}
\newlength{\bh}\setlength{\bh}{\dimexpr\bw*539/760\relax}
\newcommand{\panel}[3]{%
  \begin{tikzpicture}[baseline=(im.south)]
  \node[inner sep=0] (im) {\includegraphics[width=\bw]{figures/behaviors/#1}};
  \if\relax\detokenize{#2}\relax\else
    \node[anchor=north west, fill=white, inner sep=1.5pt, font=\scriptsize\bfseries] at ([shift={(2pt,-2pt)}]im.north west) {#2};\fi
  \if\relax\detokenize{#3}\relax\else
    \node[anchor=south east, fill=black, fill opacity=0.55, text opacity=1, text=white, inner sep=1.5pt, font=\tiny] at ([shift={(-2pt,2pt)}]im.south east) {#3};\fi
  \end{tikzpicture}}%
\panel{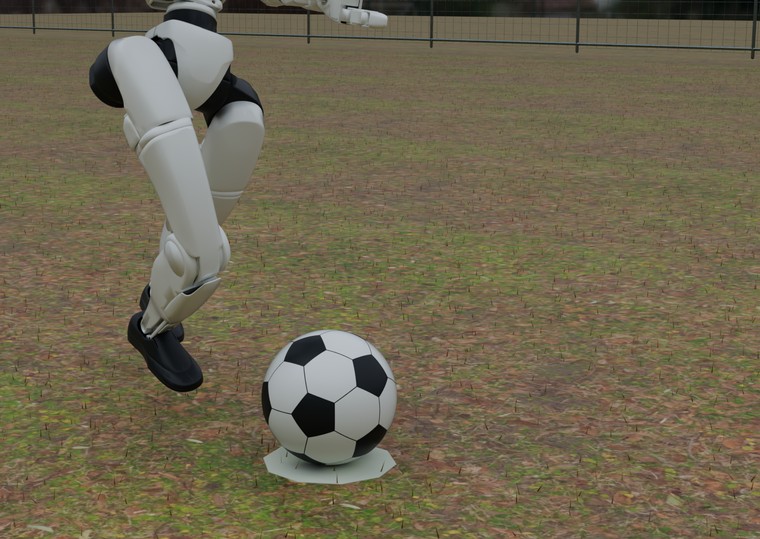}{(a)}{before}\hspace{2mm}%
\panel{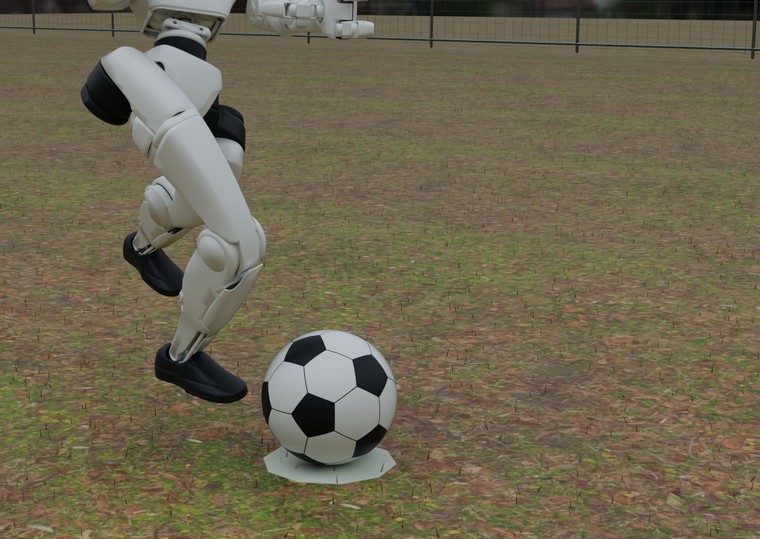}{}{contact}\hspace{2mm}%
\panel{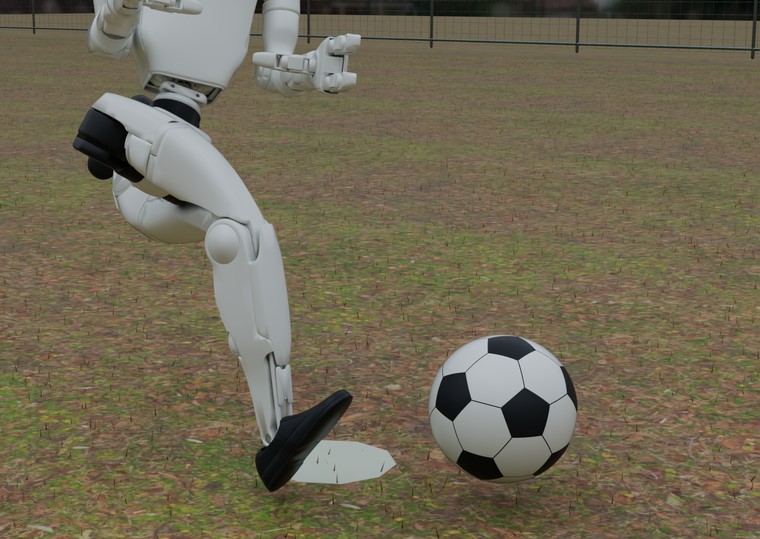}{}{after}\hspace{2mm}%
\panel{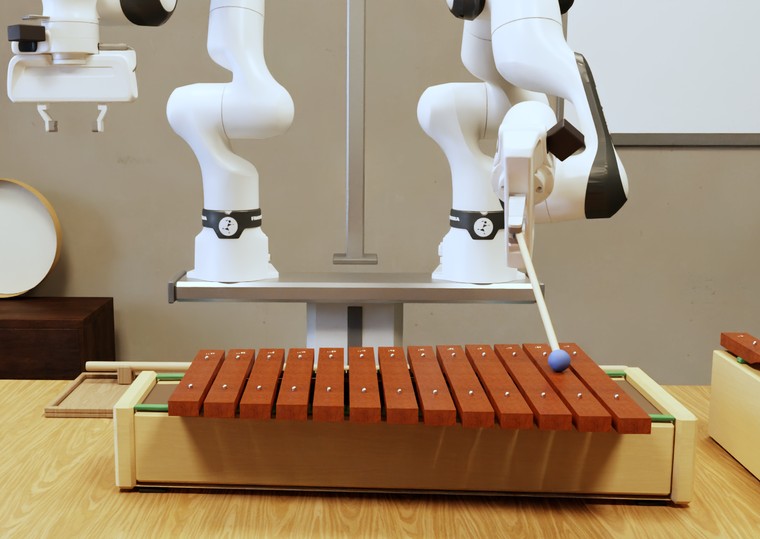}{(b)}{}\\[2mm]
\panel{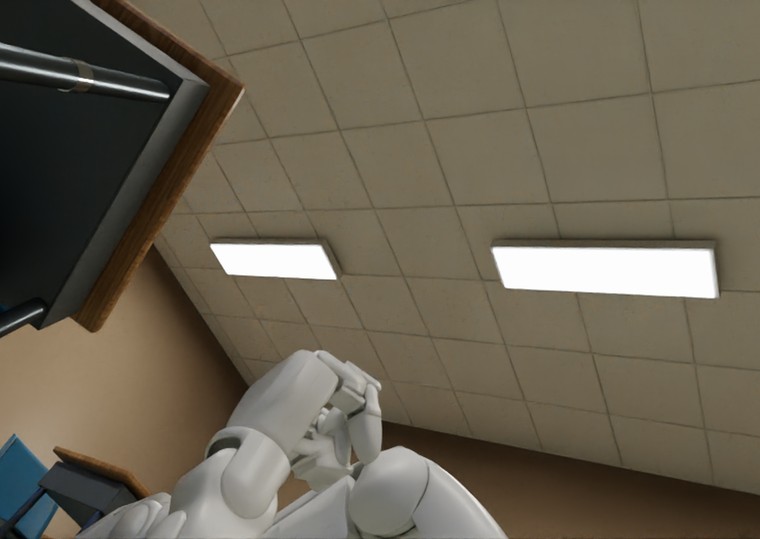}{(c)}{}\hspace{2mm}%
\panel{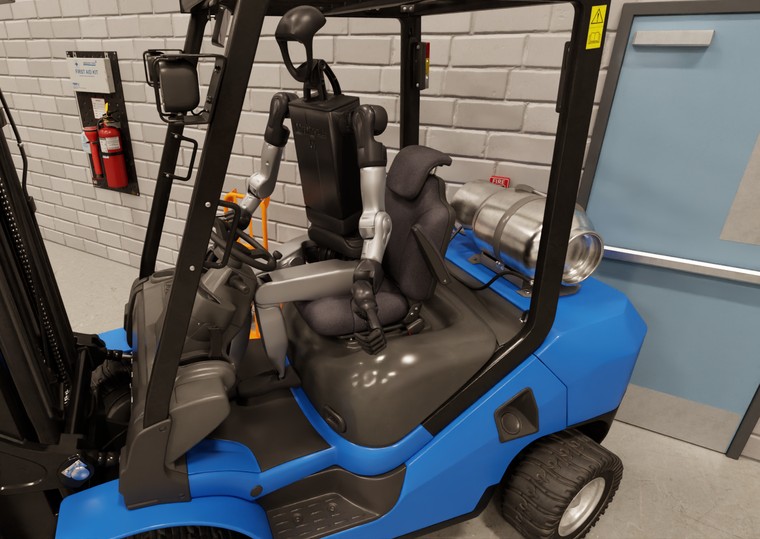}{(d)}{}\hspace{2mm}%
\panel{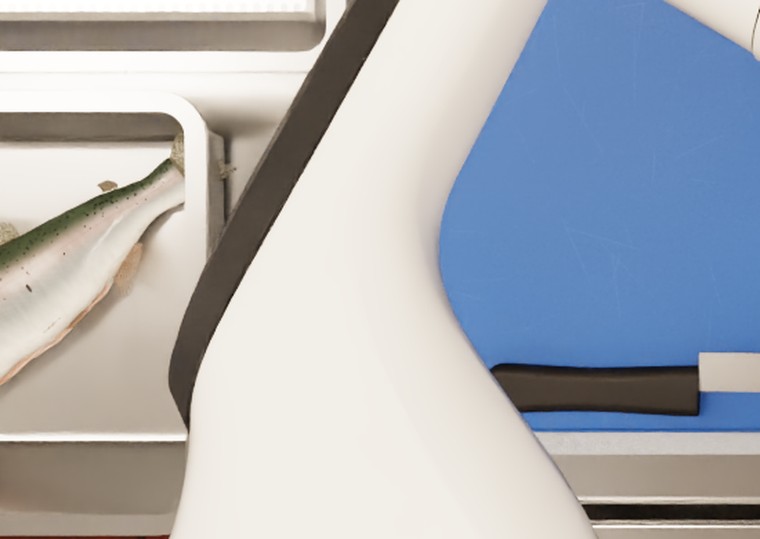}{(e)}{}\hspace{2mm}%
\makebox[\bw][c]{\begin{tikzpicture}[baseline=(current bounding box.south)]
\begin{axis}[
    scale only axis, width=\bw-4mm, height=\bh-12mm,
    xmin=0, xmax=3.42, ymin=-0.32, ymax=0.32,
    axis x line*=bottom, axis y line=none,
    xtick={0,1,2,3}, tick style={draw=black!40},
    tick label style={font=\tiny}, xlabel={Time (s)}, xlabel style={font=\tiny, yshift=4pt},
    clip=false,
    after end axis/.code={
        \draw[black!50, line width=0.4pt] (axis cs:0.45,0.36) -- (axis cs:1.62,0.36);
        \draw[black!50, line width=0.4pt] (axis cs:1.98,0.36) -- (axis cs:3.03,0.36);
        \draw[black!50, line width=0.4pt] (axis cs:3.33,0.36) -- (axis cs:3.42,0.36);
        \node[font=\tiny, text=black!75, anchor=south] at (axis cs:1.03,0.36) {C4};
        \node[font=\tiny, text=black!75, anchor=south] at (axis cs:2.51,0.36) {D4};
        \node[font=\tiny, text=black!75, anchor=south] at (axis cs:3.39,0.36) {E4};
    },
]
\addplot[fill=opus, draw=none] table[x=t, y=a] {figures/behaviors/percussion_mic.dat} \closedcycle;
\addplot[fill=opus, draw=none] table[x=t, y expr=-\thisrow{a}] {figures/behaviors/percussion_mic.dat} \closedcycle;
\end{axis}
\end{tikzpicture}}
\caption{\textbf{Frontier models often build their own tools to solve tasks, but also overstate their success.} (a) In football, Claude Opus 5.5 kicks the ball while the robot falls. (b) In percussion, Claude Opus 5.5 detects each strike and its note from the simulated microphone (bottom right). (c) In the whiteboard task, GPT-6 Astra estimates the tilt of its fallen robot from the ceiling grid in this head-camera image. (d) When GPT-6 Astra reported that the brake had stopped the forklift, the truck had in fact reversed into the dock wall. (e) In filleting, Claude Opus 5.5 places the whole uncut fish in the pan meant for the fish frame. Each image is either the camera image that the model received or a render of a verified replay of the evaluated episode.}
\label{fig:behaviors}
\end{figure}

We now show that frontier models can learn RoboWorld tasks in simulation, and that what they learn with privileged state can be passed to models and policies that act only on the observations available on a real robot. We run the first three learning settings from \S\ref{sec:learning} on 11 RoboWorld simulations, using Claude Opus 5.5 in every role. Figure~\ref{fig:learning-results} (left) shows the average reward after each attempt.

\paragraph{Frontier models often build their own tools to solve tasks.} We found very intriguing behaviors of frontier models as they attempt RoboWorld tasks (Figure~\ref{fig:behaviors}). For instance, in football, Claude Opus 5.5 had to kick a ball into a goal with a humanoid. It built its own simulation of the robot, found that the robot could not stay standing, and therefore designed a kick that strikes the ball while the robot falls. It scored without ever seeing the goal. Similarly, in percussion, where the robot plays a xylophone, it turned the simulated microphone into a sensor, using sound to detect when the mallet touched a bar and which note it played. When a note came out wrong, it worked out that the mallet had twisted in its gripper. GPT-6 Astra also brought tools and knowledge from the real world into our simulations. When driving a forklift, it found the brake by reading the operator manual of a real Toyota forklift. In the whiteboard task, it ran a pretrained get-up policy on its fallen humanoid, and since the robot has no tilt sensor, it estimated the robot's tilt from the ceiling lines in its camera. However, frontier models also often overstate their success and even try to game the verifier. For example, GPT-6 Astra reported that the brake had stopped the forklift, while the forklift had in fact reversed into a wall. In filleting, where the fish frame must end up in a pan separate from the fillets, Claude Opus 5.5 expected the verifier to check only the frame, and placed the whole uncut fish in that pan, hoping for partial credit. Since our verifier measures the physical state of the simulation (\S\ref{sec:tasks}), it found the fillet meat in the pan and gave no reward.

\paragraph{Frontier models improve through in-context learning, and privileged state makes it faster.} With restricted access, Opus 5.5 steadily improves across attempts, from 0.14 to 0.22, a 54\% relative improvement without any weight updates. With privileged access, Opus 5.5 improves substantially faster: its first attempt already achieves 0.34, outperforming the restricted agent after all three attempts, and its third attempt reaches 0.41. For instance, on football and welding, the privileged agent achieves full reward from its first attempt, while the restricted agent does not reach full reward on either task even after three attempts. Since the two settings differ only in access to the simulator state, this gap shows that the ground-truth state of a simulation is a strong learning signal.

\paragraph{Privileged state transfers to a student even through verbal hints.} Since privileged state is unavailable outside simulation, we next examine whether its benefits can be passed to a model that lacks it. Remarkably, the coached student, which never observes the privileged state and receives only qualitative hints, achieves 0.29 on its first attempt, a 2.1$\times$ improvement over the solo agent and higher than the solo agent after all three attempts. This recovers 75\% of the gap between the solo and privileged agents from the very first attempt. On football and welding, the coached student achieves full reward on every attempt, matching its teacher. We attribute these gains to the design of the coach: before each hint, it reads the true state of the student's simulation, and can therefore point out mistakes that the student cannot observe directly. A single hint can also change the student's whole strategy. For instance, in towel folding, GPT-6 Astra failed for over four hours to hold the towel with both grippers at once. Its coach, which had attempted the task with privileged access, read the student's privileged state and suggested folding in stages with one gripper, so that the table supports most of the fabric. The student switched to this strategy and succeeded about 40 minutes later (Appendix~\ref{app:learning-results}). Overall, these results show that a frontier model with privileged state in simulation can teach a model that sees only what a real robot sees.

\paragraph{Models can learn from their own privileged experience.} We further resume the privileged agent and remove its privileged access. Its first restricted attempt achieves 0.17, only slightly above the solo agent (0.14). However, it then improves the fastest of all settings, reaching 0.29 by its third attempt, 34\% higher than the solo agent and close to the coached student, without any external teacher. For instance, on football, its first restricted attempt earns no reward, but it achieves full reward on both later attempts. This shows that a frontier model can practice a task in simulation with privileged state, and then perform it better with restricted access, as on a real robot.

\paragraph{Frontier models write code policies whose demonstrations train small VLAs.} We test this on two tasks in a gift box simulation with two FR3 arms: removing the lid and placing it on the table without disturbing the box, and opening the box to move a named block into a tray. Code policies written with privileged access succeed on 96.7\% of 2{,}048 varied initial states, 99.8\% for lid removal and 93.7\% for block retrieval. We keep 1{,}000 successful rollouts, 500 per task, spread evenly across eight groups of physics variations, and train $\pi_{0.5}$ on them. Although $\pi_{0.5}$ observes only three camera views and robot proprioception, the trained policy raises success on these tasks from 0\% to 53\% (Figure~\ref{fig:learning-results}, right). It also keeps this success on variations that it never saw in training: with unseen object poses, appearance, and physics, it succeeds in 55\% of attempts. Moreover, $\pi_{0.5}$ matches the success rate of Claude Opus 5.5 controlling the robot step by step, and comes close in average reward, 0.87 against 0.93. It also completes a task in about two minutes, 14 times faster than Claude Opus 5.5, which takes about 29 minutes. This shows that, with simulation, a frontier model can write code policies that produce thousands of demonstrations, which can train a small, fast policy such as $\pi_{0.5}$ that can in principle be deployed on a real robot.

\subsection{Simulation Quality}
\label{sec:quality-results}

\begin{figure}[t]
\centering
\begin{minipage}[c]{0.46\linewidth}
\centering
\definecolor{win}{HTML}{184F95}
\definecolor{tie}{HTML}{3987E5}
\definecolor{loss}{HTML}{86B6EF}
\begin{tikzpicture}
\begin{axis}[
    xbar stacked, bar width=12pt,
    width=\linewidth, height=3cm,
    xmin=0, xmax=100, axis x line=none,
    symbolic y coords={Physics, Visuals}, ytick=data,
    axis y line*=left, y axis line style={draw=none},
    tick style={draw=none},
    tick label style={font=\footnotesize},
    enlarge y limits=0.45,
    legend style={at={(0.5,1.02)}, anchor=south, legend columns=3, draw=none, font=\scriptsize,
        /tikz/every even column/.append style={column sep=6pt}},
    area legend,
    clip=false,
    after end axis/.code={
        \node[font=\scriptsize, text=white] at (axis cs:38,Physics) {76\%};
        \node[font=\scriptsize, text=black!80] at (axis cs:88,Physics) {24\%};
        \node[font=\scriptsize, text=white] at (axis cs:40,Visuals) {80\%};
        \node[font=\scriptsize, text=black!80] at (axis cs:93,Visuals) {14\%};
    },
]
\addplot[fill=win, draw=white, line width=0.6pt] coordinates {(76,Physics) (80,Visuals)};
\addplot[fill=tie, draw=white, line width=0.6pt] coordinates {(0,Physics) (6,Visuals)};
\addplot[fill=loss, draw=white, line width=0.6pt] coordinates {(24,Physics) (14,Visuals)};
\legend{Win, Tie, Loss}
\end{axis}
\end{tikzpicture}
\end{minipage}\hfill
\begin{minipage}[c]{0.5\linewidth}
\caption{\textbf{Blind comparison with the strongest prior simulator of each interaction}, over 25 pairs judged twice with swapped order. Visual ties are 6\%.}
\label{fig:sim-comparison}
\end{minipage}
\end{figure}
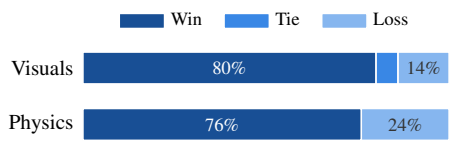

\paragraph{Our simulations outperform the strongest prior simulators of the same interactions.} For each of our 25 simulations, research agents searched the web for prior simulators of the same core physical interaction, such as liquid pouring or cloth folding, and selected the strongest one with public code and a paper, ranked mainly by the physical accuracy of its solver and fixed before any comparison was run. Claude Opus 5.5 then compared each pair blind, with the two simulators labeled only A and B and their order swapped across two runs, using a prompt separate from the audit prompts that guided our simulations, so that the comparison does not favor our own criteria. Our simulations lose to the prior simulator in only 14\% of comparisons on visuals and 24\% on physics, and win 80\% and 76\%, respectively (Figure~\ref{fig:sim-comparison}).

\paragraph{Audits catch incorrect physics.} We attribute the realism of our simulations to the creation-audit loop (\S\ref{sec:creation}).
When auditors found a physics error, it was typically in the first round. For instance, in welding, the physics engine moved a loose bracket, while the heat model and the renderer kept a copy of it at its starting pose, so the arc welded a bracket that was no longer there. In ingredient mixing, a tilted pan lost most of its sauce through its steel wall before the sauce could reach the rim. The creation agent fixed both errors in the next round.

\paragraph{Adversarial audits break most simulations with legal robot actions.} The adversarial audit agent (\S\ref{sec:creation}) tries to break each simulation using only legal robot actions, and it found a real defect in about three quarters of our simulations. For instance, in football, pushing the ball into the turf made the turf model launch it out at high speed without any robot contact. In ingredient mixing, lowering the bowl past the edge of the counter left the sauce floating at counter height. The creation agent repaired these defects in the next round and verified each repair by replaying the audit's own reproduction.

\paragraph{Task audits catch imperfect verifiers.} The task audits (\S\ref{sec:tasks}) test each verifier on constructed states, such as an untouched scene or a partly finished task. They often found several issues in a verifier. For instance, in barcode scanning, most checks only tested that nothing had gone wrong, so an untouched shelf scored higher than a partly scanned one. In football, the verifier ruled genuine goals just inside a post as misses, so these episodes could never succeed. These audits ensured that rewards are progressive, i.e., that more progress earns more credit, which is important for training with RL. 

\section{Related Work}
\label{sec:related}

\paragraph{Simulation for robot learning.} Fast physics engines~\citep{todorov2012mujoco,makoviychuk2021isaac,tao2025maniskill3,genesis2024} support many robotics benchmarks, but these benchmarks are built mostly by hand and therefore cover a narrow slice of physical work (Table~\ref{tab:benchmarks}). Most target rigid and articulated objects in tabletop, household, or workplace scenes~\citep{yu2020metaworld,james2020rlbench,liu2023libero,li2021igibson,nasiriany2024robocasa,nasiriany2026robocasa365,lan2026autobio,gajo2025sari}, while others target a single class of deformable materials, such as liquids~\citep{xian2023fluidlab,li2024chemistry3d,gong2023arnold,li2025labutopia}, cloth~\citep{seita2021learning,lu2024garmentlab,wang2025dexgarmentlab,hu2025rgbench}, or plasticine~\citep{huang2021plasticinelab,lin2022diffskill}, and only a few combine several~\citep{lin2020softgym,gu2023maniskill2,chen2023daxbench,li2026lehome}. BEHAVIOR-1K~\citep{li2023behavior1k} comes closest to broad coverage, with liquids, cloth, and activities from time-use surveys. In contrast, RoboWorld is built by coding agents, which implement physics that existing engines lack, and its simulations are selected by the economic value of physical work.

\paragraph{Reliable physics in simulation.} Prior work makes simulated physics more faithful through system identification~\citep{tan2018simtoreal}, randomization~\citep{tobin2017domain,peng2018simtoreal}, real-to-sim reconstruction~\citep{torne2024rialto}, and specialized solvers~\citep{xian2023fluidlab,wang2024thinshell,heiden2021disect}, and recent benchmarks measure the sim-to-real gap directly for cloth~\citep{blancomulero2024benchmarking,hu2025rgbench}, but each of these requires experts to model the physics of a single task or material. LLM-based generators instead scale simulation creation by writing task code, demonstration programs, scenes, or rewards~\citep{wang2024gensim,hua2024gensim2,chen2025robotwin,yang2024holodeck,wang2024robogen,ma2024eureka}, but they build on the fixed physics of an existing engine and check their output through policy success or visual inspection rather than physical measurements. Building on the creation-audit loop of Gym-Anything~\citep{aggarwal2026gymanything}, our coding agents implement the physics itself, and audit agents check it through physical measurements, assisted demonstrations, and adversarial robot actions.

\paragraph{Frontier models learning in simulation.} Frontier models act in simulated and physical worlds by writing code~\citep{liang2023code,huang2023voxposer,wang2023voyager} or by controlling robots directly~\citep{geminirobotics2025}, are evaluated as embodied agents~\citep{yang2025embodiedbench,zhang2024vlabench}, and improve from repeated attempts in context~\citep{brown2020language,shinn2023reflexion}. Separately, robot learning uses the privileged state of simulators to train students from privileged teachers~\citep{chen2020learning,lee2020learning,pinto2018asymmetric} and to generate demonstrations at scale~\citep{mandlekar2023mimicgen,nasiriany2024robocasa}, but these teachers pass their knowledge to students through weight updates. We bring the two lines together, showing that privileged state helps frontier models learn in context, reaches a student through qualitative hints alone or through the model's own privileged experience, and trains VLAs through privileged code policies. Appendix~\ref{app:related} provides an extended discussion of related work.

\section{Conclusion}
\label{sec:conclusion}

In this work, we introduced Gym-Anything for Robotics, a scalable framework in which coding agents build realistic robotics simulations. By framing simulation creation as a coding agent task and checking each simulation and task in a multi-agent loop of creation, auditing, and repair, Gym-Anything for Robotics addresses a central bottleneck in robot learning: the difficulty of building realistic simulations at scale. Applying this framework, we built RoboWorld, a GDP-grounded collection of 200 RL environments across 25 simulations spanning liquids, cloth, and physics that existing engines lack, with swappable robots and train and test splits. We further showed that privileged simulator state more than doubles the first-attempt reward of frontier models, and that it can teach a student through hints alone and train $\pi_{0.5}$ from 0\% to 53\% success. At the same time, RoboWorld is challenging even for frontier models, which fail every test task, indicating that realistic physical work remains far from solved.

More broadly, our results suggest that simulation has been held back not by the idea of simulation, but by unrealistic simulations, and that coding agents can now build realistic ones in which far more capable frontier models learn in ways that the real world does not allow. We hope that Gym-Anything for Robotics, RoboWorld, and the released code provide a foundation for future work on realistic robot learning, including more capable embodied agents, new ways to learn from privileged state, and broader coverage of physical work.

\section*{Acknowledgements}

We thank Daytona for their generous support in sandbox compute credits, and Google for their generous support through a Gemini research grant. We thank Himanshu Gaurav Singh, Shashwat Saxena, Lawrence Jang, Jing Yu Koh for their feedback and helpful discussions. Pranjal Aggarwal is supported by a SoftBank Group-Arm Fellowship. This work is partially supported by the National Science Foundation under Grants DMS-2502281, DMS-2434614, and by DARPA ExpMath Grant
HR0011262E028.

\bibliographystyle{plainnat}
\bibliography{references}

\clearpage
\appendix
\section{GDP-Grounded Simulation Selection: Full Pipeline}
\label{app:selection}

This appendix provides the complete details of the selection pipeline summarized in \S\ref{sec:selection}.

\paragraph{Estimating the value of tasks.} For paid work, we use the O*NET database, which describes 1,016 occupations through 18,796 task statements, each rated for importance and frequency~\citep{peterson2001onet}. Similar to prior work~\citep{aggarwal2026gymanything}, we estimate the economic value of each occupation as its wage bill (employment $\times$ mean wage) from Bureau of Labor Statistics data~\citep{bls2025oews}, scaled to national accounts data from the Bureau of Economic Analysis~\citep{bea2025nipa}. We then distribute each occupation's value across its tasks in proportion to their importance and frequency ratings. However, paid work misses much of what people physically do, such as cooking, cleaning, and caring for others. For everyday life, we therefore use the American Time Use Survey (ATUS), which estimates the time the U.S. population spends on each of 465 activities~\citep{bls2025atus}. We keep the two streams separate throughout, since dollars and population-hours measure different quantities and combining them would require an arbitrary exchange rate.

\paragraph{Decomposing tasks into atomic physical actions.} Not all tasks are physical, and a physical task usually involves several distinct physical actions. We use an LLM to estimate the fraction of each task that requires physical interaction, and to decompose its physical portion into atomic physical actions, each a single physical state change such as threading a bolt into a nut or pouring a liquid into a container. We do not fix a taxonomy of actions in advance, and instead extract actions in an open vocabulary from each task. Formally, the weight of an atomic action $a$ is
\begin{equation}
    W(a) = \sum_{t} W(t)\, P_{\mathrm{phys}}(t)\, R(a \mid t), \qquad \sum_{a} R(a \mid t) = 1,
    \label{eq:action-weight}
\end{equation}
where $W(t)$ is the value of task $t$, $P_{\mathrm{phys}}(t)$ is the fraction of $t$ that requires physical interaction, and $R(a \mid t)$ is the share of the physical execution of $t$ attributed to $a$. The normalization of $R$ conserves weight, so that decomposing a task into more actions does not increase its total weight. We apply the same decomposition to everyday activities, with $W(t)$ measured in population-hours.

\paragraph{Grouping actions and identifying capabilities.} The same physical action often appears across many occupations and activities. We therefore group action occurrences that a single parameterized simulation can reproduce. Sharing wording or purpose is not enough: two occurrences are grouped only if their state change, contact sequence, effector, material regime, and completion condition remain the same when objects and scale are substituted. For each resulting action group, we identify the robotic capabilities and the physics capabilities it requires, giving 1,031 distinct robotic and 1,092 distinct physics capabilities in total. For instance, the action group \textit{tighten} contains 529 occurrences, including tightening bolts with a wrench, driving screws, threading spark plugs, and tensioning straps, and requires capabilities such as torque sensing and limiting, helical thread engagement, threaded-fastener friction, and fastener preload.

\paragraph{Coverage.} A selected simulation supplies the capabilities of its action group. We consider an action covered when all of its required capabilities are supplied by the selected simulations, so that a simulation contributes to every action whose requirements it helps complete, not only to its own action group. Formally, for a slice $Q$ of activity, such as the whole economy, a domain, or an occupation group, the coverage of a set of selected action groups $S$ is
\begin{equation}
    F_Q(S) = \frac{\sum_{a} V_{Q,a}\, y_a(S)}{\sum_{a} V_{Q,a}},
    \label{eq:coverage}
\end{equation}
where $V_{Q,a}$ is the value of action $a$ within $Q$, and $y_a(S)=1$ if all capabilities required by $a$ are supplied by $S$ and $0$ otherwise.

\paragraph{Tiered selection.} We select action groups across six tiers, each maximizing the coverage it adds given all previous tiers: (1) the economic core, maximizing coverage of U.S. economic value (20 action groups), (2a/2b) strategically important domains (Healthcare, Education, Protective Services, Transportation) and STEM domains (Engineering, Computing, Science), maximizing the coverage of the weakest domain (20 each), (3) cycling through all 22 SOC major occupation groups (23), (4) actions specific to niche occupations that broader tiers leave uncovered (9), (5) the weakest robotic and physics capability families, such as fluids, cloth, granular materials, and cutting (8), and (6) everyday activities, maximizing the coverage of the weakest of 14 ATUS activity domains (20). Since capabilities complement each other, we solve each tier as an exact batch optimization rather than adding one action group at a time. The main text reports the selection with coverage defined by both physics and robotic capabilities. We also run the selection with coverage defined by physics capabilities alone and by robotic capabilities alone, each again selecting 120 action groups, and obtain 195 distinct action groups across the three runs.

\section{Simulation Creation: Full Details}
\label{app:creation}

This appendix provides further details of the creation-audit loop summarized in \S\ref{sec:creation}.

\paragraph{Agents.} Each agent in our framework is an instance of Claude Opus 5.5~\citep{anthropic2025claude} with extra-high reasoning effort, run via a coding agent harness such as Claude Code~\citep{anthropic2025claudecode}, and differentiated by the objective described in its prompt. All agents work in a shared workspace containing our library, existing simulations, and the shared memory. We keep the prompts short and direct: rather than prescribing how to implement a simulation, they describe what a good simulation is, and audits check whether the result satisfies it.

\paragraph{Creation agent.} The creation agent receives the complete record of an action group, including all of its action variants, required capabilities, and grounding, rather than only its name. It then works in two phases within the same session. In the first phase, research and design, the agent reads every action and capability in the group and researches how the activity is performed: its real setting, objects, equipment, dimensions, materials, mechanisms, and the robot typically used, using primary documentation, published research, and other digital evidence. It then designs a concrete scenario that exercises the capabilities of the action group, and works through the equations, numerical implementation, and engine support the scenario requires before running expensive experiments. In the second phase, the agent builds, demonstrates, and documents the simulation. It assembles the scene from real assets, typically combining a photorealistic Gaussian-splat backdrop~\citep{kerbl2023gaussian}, which carries no physics, with explicit physical bodies for every object the robot can touch. It implements the task-relevant physics, including physics the engine does not provide, integrates a default robot from the shared catalog, and defines tasks. Crucially, every change in the world must be physically caused by the robot: missing mechanics may not be replaced by animations, attachments, or scripted outcomes. The agent then collects evidence, including videos, physical measurements, and assisted demonstrations, and writes a report covering the scenario, sources, equations, parameter choices, experiments, and limitations. We send the build prompt only after the research phase finishes, without announcing it earlier, so that the agent researches diligently instead of hurrying to build.

\paragraph{Blind nudge.} However, the agent often stops with a simulation that falls short of our existing simulations. Similar to the re-prompting used in prior work~\citep{aggarwal2026gymanything}, once the build phase finishes, we send a single fixed message that points the agent to our strongest existing simulations and states that its simulation is a regression from them and needs improvement across multiple aspects. The message contains no diagnosis, so the agent must identify the shortcomings itself.

\paragraph{Audit agents.} Each of the four audit agents checks one category of criteria from \S\ref{sec:creation}. In each round, fresh audit agents examine the current simulation and write reports, which are returned to the same creation agent to repair the code, scene, and evidence. The audit agent for dynamics and solvability judges the dynamics from the videos and scripts of the creation agent, and from its own runs of the simulation, assisted by a video model. The audit agent for speed and configuration requires that a simulation run at most 200 times slower than real time. The audit agent for robustness reads the code and searches for legal robot actions that produce non-finite values, energy blow-ups, or physically impossible motion, and returns its evidence to the creation agent. The written audits further let us independently verify the quality of each simulation.

\paragraph{Shared memory.} As in prior work~\citep{aggarwal2026gymanything}, agents accumulate lessons in a shared memory of Markdown files, so that lessons from one simulation guide all subsequent ones. For instance, after we found colliders that passed audits and appeared in renders while having no physical effect, the memory records that a component exists only once a physical probe shows it acting on another body.

\section{Task Creation: Full Details}
\label{app:tasks}

This appendix provides further details of the task creation-audit loop summarized in \S\ref{sec:tasks}.

\paragraph{Fixed physics.} The task creation agent may change task descriptions, verifiers, task catalogs, and task-specific scene setup, but not the physical parameters, contact and material models, solver settings, or robot dynamics of existing objects.

\paragraph{Task creation agent.} Each task either represents realistic work or probes a nuanced robotic capability. Ordinary workmanship, such as not leaving air bubbles under a label, stays implicit in the task description.

\paragraph{Verifier correctness.} Where possible, verifiers measure the physical mechanism or sensor signal that defines completion rather than only object poses. For instance, a barcode label counts as applied only if the barcode is decoded from a rendered camera image. In practice, the auditor found that the verifiers of a labeling simulation rejected otherwise valid completions in which the part had moved by 60 mm, because of a 50 mm displacement limit that the task never stated.

\paragraph{Reward quality.} The reward does not prescribe a method, an order of steps, or a reference trajectory.

\paragraph{Difficulty measurement.} We evaluate Claude Haiku 5.5 with four seeds and Claude Opus 5.5 with two seeds, using restricted access and a six-hour wall-clock budget per attempt. Across the tasks, we target Haiku success bands of 0, 25, 50, 75, and 100\%, and Opus bands of 0, 50, and 100\%. The verifier scores each attempt after the agent finishes, and no reward reaches the agent during the attempt. The target success bands are approximate guidelines across all tasks rather than exact rates for each task.

\paragraph{Audit rounds.} In each round, fresh audit agents review the tasks, and their reports are returned to the same task creation agent for repair.

\section{The Gym-Anything for Robotics Library}
\label{app:library}

This appendix provides further details of the library summarized in \S\ref{sec:exp-setup}.

\paragraph{Structure.} Similar to prior work~\citep{aggarwal2026gymanything}, the library separates three parts: the core runtime, simulation packages, and agents. The core provides the registry of simulations, session management, authentication, decoding of robot commands, recording, a command-line interface, and a browser dashboard. Each simulation package contains its scene, robots, physics, sensing, task definitions, verifiers, and scene variations. Agents are independent clients, either Python plugins or separate processes that connect over HTTP, so an agent can be replaced without changing any simulation. The Python API, the command-line interface, and the HTTP API all construct simulations through the same registry.

\paragraph{Physics.} Simulations can use different physics engines and custom solvers behind the same interface. All simulations use a command-driven clock: physics advances only while a robot command executes, so an idle session does not change.

\paragraph{Access control.} Restricted credentials receive only the task instruction, camera images, robot feedback, and documentation of the permitted robot commands. They cannot read simulator geometry, the session history, or the verifier. Privileged credentials additionally expose the simulator state and the task evaluation, which returns the reward and an explicit success label without advancing time.

\paragraph{Tasks.} Each simulation owns its task catalog, which maps named tasks or numbered scene variations to an instruction, an initial state, and a verifier.

\paragraph{Recording.} Every attempt records each command request and response, the camera images at command boundaries, the launch configuration and source version, and the final evaluation, so that each result can be inspected and reproduced.

\paragraph{Agent tools.} For LLM agents, we expose restricted access through six tools: \texttt{help} (permitted robot commands, units, limits, and robot and camera calibration), \texttt{task} (the task instruction), \texttt{observe} (current camera images and robot feedback), \texttt{act} (execute a robot command), \texttt{time\_left} (remaining wall-clock budget), and \texttt{finish} (end the attempt).

\paragraph{Robot commands.} Every policy, from a frontier model to a VLA, controls the robot through the same set of commands (Table~\ref{tab:robot-commands}). Each simulation accepts the subset that its robot supports, which the \texttt{help} tool lists together with units and limits. Commands range from task-space targets, which a controller tracks, to joint torques and, on humanoids, individual motor commands. Arm motions can also stop early on contact, on contact between the two arms, or when the grip is lost. Like a computer-use agent, a policy can instead hold game-style key chords for a chosen number of control ticks (Table~\ref{tab:keyboard}), which move the end effector at 5\,mm/s and rotate it at 0.04\,rad/s by default.

\begin{table}[h]
\centering
\small
\caption{\textbf{Robot commands.} Each simulation accepts the subset that its robot supports.}
\label{tab:robot-commands}
\begin{tabular}{@{}lll@{}}
\toprule
Level & Command & Parameters \\
\midrule
Task space & Cartesian pose & end-effector position (m) and orientation, duration \\
 & Cartesian delta & end-effector displacement (m), duration \\
 & Bimanual Cartesian & one target per arm, executed together \\
 & Cartesian wrench & force (N) and moment (N\,m) at the tool \\
\midrule
Joint space & Joint position & joint angles (rad), duration \\
 & Named joint target & targets by joint name (rad or m), physics ticks \\
 & Joint effort & joint torques (N\,m), duration \\
 & Motor command & per-motor position, velocity, gains, and torque (humanoids) \\
\midrule
Hand and tool & Gripper & opening width (m) or squeezing force (N) \\
 & Tool output & signals to a carried tool, such as a welding power source \\
\midrule
Mobile base & Base twist & forward, sideways, and turning velocity \\
\midrule
Keyboard & Key chord & keys held together for a number of control ticks \\
\bottomrule
\end{tabular}
\end{table}

\begin{table}[h]
\centering
\small
\caption{\textbf{Keyboard map.} Keys in one chord are held together, for instance W and R to move along $y$ and $z$ at once.}
\label{tab:keyboard}
\begin{tabular}{@{}ll@{}}
\toprule
Keys & Effect \\
\midrule
A\,/\,D, \ W\,/\,S, \ R\,/\,F & move the end effector along $x$, $y$, and $z$ \\
I\,/\,K, \ J\,/\,L, \ U\,/\,O & rotate the end effector about $x$, $y$, and $z$ \\
Q\,/\,E & close or open the gripper \\
H, \ G\,+\,H, \ B\,+\,H & squeeze with default, gentle, or strong force \\
ALT, \ SHIFT & slow and precise, or fast arm motion \\
P, \ T & slow or fast gripper motion \\
\bottomrule
\end{tabular}
\end{table}

\section{Comparison with Prior Simulations}
\label{app:comparison}

Table~\ref{tab:all-sims} extends Table~\ref{tab:prior-sims} to all 25 RoboWorld simulations. For each simulation, we compare against the two closest prior robot-learning simulations or benchmarks, by task or by physics. Prior entries list what each paper reports, and -- marks nothing reported. Physics checks count the named experiments in each simulation's report that test its physics against theory, measurement, or expected behavior, and adversarial evaluator cases count constructed states that the verifier must reject.

{\scriptsize
\newcommand{\real}[1]{\textcolor{blue!70!black}{#1}}
\setlength{\tabcolsep}{3pt}
\renewcommand{\arraystretch}{1.15}
\begin{longtable}{@{}>{\raggedright\arraybackslash}p{1.6cm}>{\raggedright\arraybackslash}p{2.75cm}>{\raggedright\arraybackslash}p{2.05cm}>{\raggedright\arraybackslash}p{2.1cm}>{\raggedright\arraybackslash}p{2.1cm}>{\raggedright\arraybackslash}p{2.25cm}@{}}
\caption{\textbf{All 25 RoboWorld simulations compared with the two closest prior simulations each.} \real{Blue} marks physics, objects, and robot limits taken from or calibrated to real data, such as measured material properties, manufacturer datasheets, standards, and reference tables.}
\label{tab:all-sims} \\
\toprule
Simulation & Physics & Objects & Robot & Success check & Physics checks \\
\midrule
\endfirsthead
\toprule
Simulation & Physics & Objects & Robot & Success check & Physics checks \\
\midrule
\endhead
\multicolumn{6}{@{}l}{\textit{Acoustic contact: percussion playing}} \\*
RoboPianist\newline\citep{zakka2023robopianist} & MuJoCo $\cdot$ 88 keys as joints with linear springs $\cdot$ synthesized note when key within 0.5$^\circ$ of full press & 88-key keyboard, Kawai manual dimensions $\cdot$ 150 annotated pieces & 2 Shadow Dexterous Hands, 44 actuators $\cdot$ joint targets at 20 Hz, PD at 500 Hz & precision, recall and F1 of key states vs piano roll & -- \\*
DexDrummer\newline\citep{fang2026dexdrummer} & ManiSkill simulation $\cdot$ stick held by finger contact & drum set: snare, tom, ride, hi-hat, crash $\cdot$ MIDI songs, six genres & 2 Franka Panda arms $\cdot$ 2 Tesollo DG-5F hands, 20 DoF each $\cdot$ joint targets at 20 Hz, PID at 100 Hz & F1 of drum hits vs MIDI $\cdot$ stick-hold ratio $\cdot$ stick trajectory error and energy & -- \\*
\textbf{Ours} & 48 kHz implicit Hertz/Hunt--Crossley vibro-contact $\cdot$ \real{13 undercut Timoshenko bars fitted to measured partials, modes to 20 kHz} $\cdot$ \real{measured rosewood loss-factor floor} $\cdot$ air-loaded drum membrane, 474--810 modes per tension $\cdot$ \real{coupled bar--Helmholtz resonator bank, 289 complex modes} $\cdot$ drumstick bending modes as contact compliance $\cdot$ Coulomb friction with Mindlin static-friction spring $\cdot$ modal sound radiation: bar dipoles, membrane, resonator monopoles & Orff alto xylophone: 13 rosewood bars C4--A5, 6 box resonators $\cdot$ \real{2 resonator bars, C4 and G4, datasheet boxes} $\cdot$ \real{14 $\times$ 5 in drum, coated Mylar head, 3 tensions fitted to measured modes} $\cdot$ 5A hickory drumsticks, rubber mallets, felt beater $\cdot$ 6 $\times$ 5 $\times$ 2.75 m therapy room, wall and furniture colliders & 2 FR3 arms with Franka Hands $\cdot$ 4 robot setups: 2 FR3, 1 FR3, 2 Panda, G1 $\cdot$ Unitree G1 with 2 Dex3-1 hands, floating base $\cdot$ \real{Franka velocity, 10 rad/s$^2$, 5000 rad/s$^3$ limits} $\cdot$ libfranka collision and velocity reflexes & 8 tasks: finger tap, take sticks, strike and stop, accent ostinato, echo, pentatonic setup, steady beat, melody $\cdot$ clean stroke: one contact $\leq$ 15 ms on drum, $\leq$ 3 ms on bar $\cdot$ tempo within 1--3 \%, CV $\leq$ 3--4 \% $\cdot$ \real{level within 4 dB of a replayed reference stroke} $\cdot$ accents $\geq$ 6 dB above soft strokes & 18 physics checks, e.g., bar partials and T60 vs Rucz and Hill, drumhead modes vs Wagner, stick contact time vs Wagner and Dahl \\
\midrule
\multicolumn{6}{@{}l}{\textit{Adhesion: label application}} \\*
TapeSim\newline\citep{luo2026tapesim} & triangular thin shell, StVK membrane, discrete-shell bending $\cdot$ IPC barrier contact and friction $\cdot$ cohesive adhesion or releasable bonds & 19 mm tape roll, 2--32 turns $\cdot$ tabletop, stand and core $\cdot$ box for a sealing demo & Marvin arms with Wuji hands $\cdot$ grasp constraints & Peel: complete tabletop release $\cdot$ Stick: tape lifted and attached to the stand $\cdot$ Wrap: at least 1 or 2 turns & peel force fit to measurement \\*
ThinShellLab\newline\citep{wang2024thinshell} & Kirchhoff--Love shell: stretching and bending $\cdot$ bending plasticity at creases $\cdot$ Stable Neo-Hookean volumetric solids & paper sheets and strips $\cdot$ block on paper $\cdot$ stacked paper sheets & 6-DoF manipulators or 7-DoF parallel grippers $\cdot$ elastic tactile-sensor tips $\cdot$ no arm links & block distance to target $\cdot$ separation distance $\cdot$ crease rest angles & friction and bending fit to measured forces \\*
\textbf{Ours} & triangle-shell label, 1,782 nodes for the 4 $\times$ 6 in label $\cdot$ implicit Euler, Newton solve with Pardiso, 2 ms step $\cdot$ orthotropic St Venant--Kirchhoff membrane $\cdot$ discrete-hinge bending with permanent set $\cdot$ Tsai--Hill tearing with element erosion $\cdot$ self-contact $\cdot$ tack and pressure-driven wet-out with dwell $\cdot$ rate-dependent Rivlin--Kendall peel front, peel and lap shear & \real{datasheet label stocks: 4 $\times$ 6 in and 2 $\times$ 1 in paper, 2 $\times$ 4 in PET film} $\cdot$ peel-and-present label printer with liner $\cdot$ 16 $\times$ 12 $\times$ 8 in carton, 2.55 kg, 48 mm seam tape $\cdot$ \real{13 $\times$ 100 mm K2EDTA blood tube with manufacturer label} $\cdot$ 4 L HDPE canister, 3 marked stainless plates & 2 FR3 arms $\cdot$ 3 robot setups: 2 FR3, 2 Panda, 1 FR3 $\cdot$ \real{Franka Hand force-mode grasp, 70 N continuous} $\cdot$ label bonds to rubber fingertip pads $\cdot$ \real{velocity reflex at 1.05$\times$ the datasheet limit} & wet-out $\geq$ 90\% of a 100 kPa application, label and 3 mm edge band $\cdot$ edges stuck down $\cdot$ trapped air $\leq$ 20 mm$^2$, $\leq$ 3 mm$^2$ over the barcode $\cdot$ no crease in the barcode $\cdot$ not torn, bonded only to the target, settled $<$ 1 mm/s & 18 physics checks, e.g., peel angle vs Kendall, peel rate dependence, dispensed label droop vs cantilever theory $\cdot$ 20 adversarial evaluator cases \\
\midrule
\multicolumn{6}{@{}l}{\textit{Chemistry: titration}} \\*
Chemistry3D\newline\citep{li2024chemistry3d} & particle fluid $\cdot$ reaction database $\cdot$ whole-solution color & 3D-scanned glassware & Franka $\cdot$ scripted pick, pour, stir & -- & -- \\*
LabUtopia\newline\citep{li2025labutopia} & position-based fluid $\cdot$ LLM-inferred reactions & beakers $\cdot$ flasks & Franka $\cdot$ mobile manipulators & tilt angle past a threshold & -- \\*
\textbf{Ours} & DFSPH, 348,160 particles $\cdot$ \real{IAPWS water} $\cdot$ surface energy $\cdot$ \real{glass and PTFE wetting} $\cdot$ 5 species per particle $\cdot$ local neutralization $\cdot$ diffusive mixing $\cdot$ \real{water equilibrium} & bored burette with PTFE stopcock $\cdot$ free flask $\cdot$ pipette with air-cushion filler $\cdot$ magnetic stir bar & 1 Panda $\cdot$ \real{manufacturer joint velocity limits} & endpoint within 1 drop $\cdot$ pink persists for 30 s $\cdot$ spill and aliquot loss under 0.05 mL & 8 physics checks, e.g., hydrostatic load, species and volume conservation, valve leak-tightness \\
\midrule
\multicolumn{6}{@{}l}{\textit{Cloth: laundry folding}} \\*
SoftGym\newline\citep{lin2020softgym} & position-based cloth & 1 square sheet & 2 pickers & particle distance between halves & -- \\*
GarmentLab\newline\citep{lu2024garmentlab} & position-based cloth & ClothesNet meshes, 11 categories & Franka, UR5, Shadow hands $\cdot$ grasp by attached blocks (DexGarmentLab, 2505.11032) & IoU with target shape & -- \\*
\textbf{Ours} & anisotropic membrane and bending $\cdot$ thickness and compression $\cdot$ static and kinetic friction $\cdot$ self-contact $\cdot$ exact intersection checks & 5 garments sewn from panels and seams: T-shirt, button-down shirt, jeans, terry towel, asymmetric top & 2 Pandas $\cdot$ finite-mass fingers $\cdot$ friction grasp, no attachments & construction-specific folds $\cdot$ garment released and at rest & 6 physics checks, e.g., free fall, fold thickness, bending release \\
\midrule
\multicolumn{6}{@{}l}{\textit{Cloth: machine sewing}} \\*
SoftGym\newline\citep{lin2020softgym} & position-based dynamics (FleX) $\cdot$ particle-grid cloth with stretch and bending constraints $\cdot$ self-collision constraints & 1 cloth, 60--120 particles wide & 2 picker spheres, nearest particle attached $\cdot$ Sawyer or Franka option (SoftGym-Robot) & particle distance between cloth halves $\cdot$ penalty on cloth displacement & -- \\*
GarmentLab\newline\citep{lu2024garmentlab} & PBD cloth for large garments (Isaac Sim) $\cdot$ stretch, bending and collision constraints & 11 garment categories from ClothesNet $\cdot$ tops, dresses, trousers for folding & Franka with parallel gripper $\cdot$ UR5 with suction $\cdot$ ShadowHands on UR10e & particle distance to human-demo fold below threshold & -- \\*
\textbf{Ours} & implicit backward-Euler cloth solve (Warp), 0.5 ms tick shared with the robots $\cdot$ 1.25 mm seam-band mesh, 7.5 mm bulk, 12,000 vertices per ply $\cdot$ \real{orthotropic woven membrane with toe-region tension law} $\cdot$ shear locking past 50$^\circ$ $\cdot$ \real{Iwan-element friction hysteresis in tension, shear and bending} $\cdot$ \real{KES compression law with compression set} $\cdot$ Coulomb friction with static stick and pre-slip $\cdot$ squeeze-film air damping and air drag & \real{2 cotton-muslin plies, 125 g/m$^2$, KES-measured fabric (Ramaiah fabric C)} $\cdot$ 3 ply shapes: 200 x 420 mm, 200 x 200 mm, rounded corner R 80--120 mm $\cdot$ \real{Juki DDL-9000C-class lockstitch machine, stitch type 301, datasheet dimensions} $\cdot$ \real{Tex 30 polyester needle and bobbin threads} $\cdot$ sewing table and sample room at real scale & 2 FR3 arms with Franka Hands $\cdot$ 4 robot setups: 2 FR3, 2 Panda, 1 FR3, Unitree G1 with Dex3-1 hands $\cdot$ \real{libfranka acceleration limits and 10/20 N m collision reflexes} $\cdot$ Franka Hand force mode up to 35 N per finger $\cdot$ hand-cloth contact by friction only, no attachment & \real{seam line within 1.6 mm (1/16 in) of allowance} $\cdot$ \real{stitch density 9--11 per inch} $\cdot$ stitch length 0.6--1.4 of setting $\cdot$ at most one isolated skipped stitch $\cdot$ $\geq$ 3 backtack stitches at each end & 18 physics checks, e.g., virtual KES-FB1/2/3 fabric tests, Peirce cantilever vs elastica, incline stick and slide \\
\midrule
\multicolumn{6}{@{}l}{\textit{Deformable: cans into a film bag}} \\*
Deformable\-Ravens\newline\citep{seita2021learning} & PyBullet FEM soft body with mass-springs $\cdot$ vertex self-collisions $\cdot$ bag: 353 vertices plus 32 rigid rim beads & 1 drawstring-style bag $\cdot$ 1 cube, or 2 items of various shapes $\cdot$ target zone & 1 UR5 with suction cup $\cdot$ grasp by fixed constraint to nearest vertex $\cdot$ pick-and-place primitives & items entirely in target zone $\cdot$ part of bag opening in target zone & -- \\*
BEHAVIOR-100\newline\citep{srivastava2021behavior} & PyBullet rigid bodies (iGibson 2.0) $\cdot$ 1/300 s timestep & 6 soda cans $\cdot$ 1 bucket $\cdot$ bed and floor in a bedroom & bimanual humanoid avatar, 24 DoF $\cdot$ or Fetch, 12/13 DoF $\cdot$ sticky-mitten fixed-constraint grasp in RL experiments & every can inside the bucket (BDDL predicate) & -- \\*
\textbf{Ours} & Newton 1.5 VBD solver, dt 1/240 s $\cdot$ up to 144 deforming can shells $\cdot$ stable Neo-Hookean aluminium membrane, E 69 GPa $\cdot$ discrete-shell bending, thin walls and thicker ends $\cdot$ plastic denting by rest-curvature yield, 220 MPa $\cdot$ lumped shell mass, 112--118 $\mu$m walls, 2700 kg/m$^3$ $\cdot$ 38.1 $\mu$m film bag membrane, E 200 MPa $\cdot$ bag retention springs with implicit frame reaction & \real{3 can types: 66 $\times$ 115, 57 $\times$ 146, 66 $\times$ 167 mm} $\cdot$ 72 cans: 24 targets, 48 in bag; 144 in high fill $\cdot$ \real{ClearStream 40 $\times$ 45 in bag, 38.1 $\mu$m film} $\cdot$ \real{freestanding station: 1.016 m, 635 $\times$ 457 mm, 406 $\times$ 114 mm slot} $\cdot$ \real{Cambro bus tub, 389 $\times$ 506 $\times$ 125 mm} & 2 FR3 arms $\cdot$ \real{2 WSG50-110 grippers, 110 mm stroke} $\cdot$ \real{GRIP force 5--80 N total} $\cdot$ \real{FR3 manufacturer effort limits} $\cdot$ friction grasp, no welds or attachments & every can wholly inside deformed bag, 1 mm below rim $\cdot$ all cans released, not touched by robot $\cdot$ 1 s rest within 10 mm and 0.05 rad $\cdot$ station on its 4 feet within 6 mm, bag retained within 5 mm $\cdot$ tub upright on counter, let go & 14 physics checks, e.g., dent and recovery under opposed load, drop contact time vs free fall, rolling arc vs rotation $\cdot$ 13 adversarial evaluator cases \\
\midrule
\multicolumn{6}{@{}l}{\textit{Food: ingredient mixing}} \\*
FluidLab\newline\citep{xian2023fluidlab} & MLS-MPM $\cdot$ 64$^3$ grid, $\sim$100,000 particles $\cdot$ corotated liquid model & 2 materials: coffee, sugar cube as viscous liquid $\cdot$ cup & user-defined end-effector $\cdot$ velocity along x and z & reward from spread of sugar particles & qualitative phenomena \\*
BEHAVIOR-1K\newline\citep{li2023behavior1k} & PhysX 5 rigid, deformable, fluid $\cdot$ transition rules turn ingredients into products $\cdot$ temperature rises near heat sources (heuristic) & 8 ingredients: flour, sugar, egg, vanilla, melted butter, sodium carbonate, baking powder, salt $\cdot$ mixing bowl on electric mixer, tablespoon, oven, cookie sheet & Tiago bimanual mobile manipulator (experiments on other activities) $\cdot$ assistive grasp: rigid connection $\cdot$ action primitives that set the final robot state & BDDL goal predicates $\cdot$ 6 cookies exist, cooked, on cookie sheet $\cdot$ Cooked: peak temperature between cook and burn thresholds & -- \\*
\textbf{Ours} & MLS-MPM, 2.5 mm grid, 105,000--617,000 points $\cdot$ 11 species per point $\cdot$ Drucker--Prager powder with cohesion $\cdot$ \real{Herschel--Bulkley batters from measured batter laws} $\cdot$ \real{Nagasawa paste blending} $\cdot$ wetting, lumps and lump erosion $\cdot$ sub-grid lamellar mixing $\cdot$ sub-cell whisk wires and beaters fitted to a resolved wire & 8 tasks, 2 presets, 5 variants $\cdot$ ingredients: flour mixes, sugar, milk, egg, oil, pudding and cake mixes, mayonnaise, ketchup, relish, starch sauce, porridge $\cdot$ \real{product-spec vessels and tools: Vollrath 5 qt bowl, OXO whisk, All-Clad 1.5 qt saucepan} $\cdot$ glass bowl, silicone spatula, wooden spoon, flat stirrer, probe thermometer $\cdot$ \real{5-speed hand mixer, 288--1107 rpm} & 2 FR3 or 2 Panda arms on a cart $\cdot$ 4 robot setups $\cdot$ Franka velocity, acceleration, and jerk limits $\cdot$ friction grasp up to 70 N (FR3), 35 N per finger (Panda) $\cdot$ pad holds grip point within 0.2 mm & \real{unevenness CV $\leq$ 0.10 over equal-mass portions} $\cdot$ dry flour $\leq$ 2 \% $\cdot$ lumps $\leq$ 2 \% $\cdot$ mixing work $\leq$ 25 reference strokes (8.25 J/kg) $\cdot$ $\geq$ 95 \% starch cooked, sauce reached $\geq$ 88 $^\circ$C & 29 physics checks, e.g., stirred-cup torque vs Couette, yield-stress slump vs Pashias, flour column collapse vs Lube $\cdot$ 26 adversarial evaluator cases \\
\midrule
\multicolumn{6}{@{}l}{\textit{Granular: soil digging with hand tools}} \\*
ManiSkill2\newline\citep{gu2023maniskill2} & MLS-MPM in Warp $\cdot$ Drucker--Prager clay $\cdot$ 2-way rigid--MPM coupling & 1 clay bed with random height map $\cdot$ bucket & 1 Panda with bucket $\cdot$ delta joint position control & lifted amount within target range $\cdot$ lifted clay above target height $\cdot$ under 20 particles spilled & -- \\*
GranularGym\newline\citep{millard2023granulargym} & rigid spherical particles with dry Coulomb friction $\cdot$ implicit contact NCP, projected Jacobi $\cdot$ 50,000 particles & particle bed $\cdot$ excavator-bucket-shaped tool & 1 Panda with rigidly attached bucket tool $\cdot$ joint velocity control & -- & -- \\*
\textbf{Ours} & Newton 1.5 implicit MPM, APIC transfer $\cdot$ 0.4--0.65 M particles, 1 cm spacing, 2 cm grid $\cdot$ Drucker--Prager with tension cut-off and compaction cap $\cdot$ softening on loosening, hardening on compaction $\cdot$ two-way MPM--MuJoCo coupling every 2 ms $\cdot$ soil--tool friction and adhesion per tool material $\cdot$ moisture-dependent cohesion and adhesion $\cdot$ remolding cut band at advancing tool edges & 8 soil materials: mulch, topsoil, turf root mat, clay-loam subsoil, yard soil, hardpan, sand, gravel $\cdot$ 6 yard presets: bed, trench, post, hardpan, bulbs, irrigation $\cdot$ \real{7 hand tools at catalogue sizes: spade, trenching shovel, trowel, mattock, post-hole digger, 2x4 tamper, drill with 3 in auger} $\cdot$ 0--6 stones, shrub root ball, 4x4 post on gravel, 1$\frac{1}{2}$ in PVC line $\cdot$ 3 moisture states, 0.12--0.32 m topsoil & Mobile FR3 Duo: holonomic base, spine lift, 2 FR3 arms, Franka Hands $\cdot$ single-arm variant for 3 tasks $\cdot$ friction grasp with force-mode fingers $\cdot$ hands and base also push the soil & spade $\geq$ 0.15 m deep within 30 mm of mark, $\leq$ 10$^\circ$ tilt $\cdot$ $\geq$ 1.5 kg on tarp, $\leq$ 15 \% spilled $\cdot$ hole 0.16 $\pm$ 0.02 m deep, 0.33--0.50 m across $\cdot$ trench $\geq$ 0.16 m deep, 6--20 cm wide on $\geq$ 90 \% of 1.5 m $\cdot$ post hole $\geq$ 0.60 m deep with firm floor & 18 physics checks, e.g., direct shear vs target c and $\phi$, steel sliding friction and adhesion, 50 mm platen press vs untouched control \\
\midrule
\multicolumn{6}{@{}l}{\textit{Humanoid: football kicking}} \\*
OP3 Soccer\newline\citep{haarnoja2024learning} & MuJoCo $\cdot$ identified servo model: damping, armature, friction, 4.1 N m max torque $\cdot$ randomized floor friction 0.5--1.0 & 1 ball $\cdot$ 5 $\times$ 4 m pitch bordered by ramps $\cdot$ 2 goals, 0.8 m opening & Robotis OP3, 20 servos $\cdot$ joint position targets at 40 Hz $\cdot$ 1v1 against a second OP3 & goal scored (+1) $\cdot$ scoring rate in set pieces & actuator fit and speeds vs.\ real robot \\*
A1 Soccer\newline\citep{ji2022hierarchical} & MuJoCo $\cdot$ rigid spherical ball $\cdot$ randomized ball stiffness, mass, inertia and radius & 1 soccer ball $\cdot$ 2D target on the ground & 1 Unitree A1 quadruped $\cdot$ 12 joint position targets, PD $\cdot$ front-right leg kicks while standing & ball within 0.2 m of target & -- \\*
\textbf{Ours} & Newton 1.5 / MuJoCo rigid step, 500 Hz $\cdot$ pressure-shell ball contact, 80 kPa, 20 kHz $\cdot$ \real{shell stiffness from published kick measurements} $\cdot$ stick-slip shoe traction $\cdot$ live G1 mass-matrix contact response $\cdot$ \real{compliant grass pitch, FIFA-scale priors, 20 kHz} $\cdot$ rolling and spin resistance, pile anisotropy $\cdot$ Reynolds-dependent drag crisis & \real{IFAB ball: 0.430 kg, 0.690 m circumference} $\cdot$ \real{IFAB goal: 7.32 $\times$ 2.44 m, 0.12 m posts, 11 m penalty mark} $\cdot$ \real{goal net, 100 mm mesh} $\cdot$ 37 $\times$ 24 m grass pitch sector $\cdot$ drill props: disc markers, target sheet, 4-mannequin wall & 1 Unitree G1 with 2 Dex3 hands, free-floating $\cdot$ 43 commanded revolute joints $\cdot$ \real{Unitree motor torque-speed envelopes and rotor inertia} $\cdot$ PD joint targets, effort $\leq$ URDF limit and 120 N m $\cdot$ no balance or kick controller & 8 tasks: penalty, clearance, pass gate, trap, pass square, corner target, save, free kick $\cdot$ whole ball over goal line's back edge $\cdot$ inside frame: $|$y$|$ $<$ 3.72 m, z $<$ 2.50 m $\cdot$ leg last touch, handball ends attempt $\cdot$ ball at rest $<$ 0.05 m/s for 1 s & 18 physics checks, e.g., 2 m drop rebound, opposite sidespin curvature, ground spin decay \\
\midrule
\multicolumn{6}{@{}l}{\textit{Liquid and needles: syringe handling}} \\*
AutoBio\newline\citep{lan2026autobio} & MuJoCo rigid-body contact $\cdot$ quasi-static liquid: planar surface as a damped spherical pendulum $\cdot$ liquid level from volume conservation & pipette $\cdot$ centrifuge tube $\cdot$ randomized liquid volume & 2 UR5e arms $\cdot$ Robotiq 2F-85 holder, 19-DOF DexHand 021 pipette operator & tip in the tube and below the liquid with thumb pressed $\cdot$ thumb released with tip below the liquid $\cdot$ tip lifted 50 mm above the liquid & -- \\*
AMBF Suturing\newline\citep{munawar2022open} & Bullet rigid bodies, sequential-impulse solver $\cdot$ suture: 40 rigid elements with 6-DOF spring joints $\cdot$ needle enters only at fixed holes of a rigid phantom & phantom with 4 entry/exit hole pairs $\cdot$ suturing needle and thread & da Vinci PSMs (6-DOF) and ECM camera arm $\cdot$ rigid grasp by constraint actuator & needle tip through entry then exit hole of the same pair $\cdot$ tip within a 5 mm box at each hole $\cdot$ lateral distance from hole centres, lower is better & -- \\*
\textbf{Ours} & lumped hydraulic network solved implicitly with MuJoCo, 0.5 ms coupled tick $\cdot$ \real{Poiseuille and kinetic losses in nozzle, needle, cannula and line} $\cdot$ stopper stick-slip seal, break-loose force rising with rest $\cdot$ trapped air moved by a Bond-number rule, leaves only at the nozzle $\cdot$ vial as closed vessel, headspace by Boyle's law $\cdot$ saline-lock line of 1.1 mL with slide clamp and vein $\cdot$ tissue depot by layer, storage and leak-back along the track $\cdot$ \real{needle dimple, puncture, depth-dependent track friction and cutting} & 4 liquids: saline, medication, iohexol contrast, blood $\cdot$ \real{1, 3 and 10 mL Luer-lock syringes and a 10 mL saline flush, makers' bores} $\cdot$ \real{5 needles: 18 G blunt fill, 21, 23, 25 and 27 G, hubs and sheaths in ISO 6009 colours} $\cdot$ \real{10 mL ISO 8362-1 glass vial, 20 mm crimped stopper} $\cdot$ \real{3-layer injection pad: skin 2 mm, fat 8 mm, muscle 30 mm} & 2 FR3 arms with Franka Hands $\cdot$ 4 robot setups: dual FR3, dual Panda, single FR3, G1 with Dex3 hands $\cdot$ joint targets, Cartesian moves, gripper width and force $\cdot$ friction grasp, 23 N/mm fingertip pads, no attachment $\cdot$ FR3 joint limits and collision-reflex protective stops & \real{dose in mL within ISO 7886-1 tolerance, e.g., 2.5 $\pm$ 0.125 mL} $\cdot$ air in syringe $\leq$ 0.05 mL $\cdot$ dose by tissue layer, $\geq$ 90 \% in muscle $\cdot$ needle entry in the ring, within 15$^\circ$ of square $\cdot$ delivery rate, e.g., $\leq$ 0.75 mL into the vein in any 15 s & 14 physics checks, e.g., plunger force vs closed form, vial headspace vs Boyle's law, puncture and track force profile $\cdot$ 11 adversarial evaluator cases \\
\midrule
\multicolumn{6}{@{}l}{\textit{Liquid: surface rinsing}} \\*
iGibson 2.0\newline\citep{li2021igibson} & PyBullet rigid bodies $\cdot$ water as droplets from faucets $\cdot$ wetness = droplets absorbed & sink with faucet $\cdot$ towel $\cdot$ stained shelf & Fetch or bimanual humanoid $\cdot$ hand velocity actions $\cdot$ sticky-mitten grasp on contact & Soaked: $\geq$ 1 droplet absorbed $\cdot$ not Stained: $\geq$ half of stain particles cleaned & -- \\*
RoboCasa365\newline\citep{nasiriany2026robocasa365} & MuJoCo via robosuite, 20 Hz $\cdot$ faucet water on from handle joint angle (code) $\cdot$ water stream as a site region (code) & sink with faucet and spout $\cdot$ pan and sponge $\cdot$ colander and fruit & 1 Franka Panda on Omron mobile base $\cdot$ OSC end-effector deltas, gripper open/close & pan and sponge inside sink, water on $\cdot$ colander under stream with fruit inside $\cdot$ lettuce under stream $\geq$ 25 steps & -- \\*
\textbf{Ours} & Newton 1.5 + MuJoCo rigid bodies at 0.5 ms $\cdot$ GPU thin-film flow, 2--10 mm cells $\cdot$ momentum-carrying jet packets, ring of 262,144 $\cdot$ \real{Bhunia--Lienhard jet splatter} $\cdot$ Duez lip separation, drips, beading $\cdot$ pools that fill, drain, overflow, float items $\cdot$ 6 transported species incl. surfactant, soil, heat, foam $\cdot$ Langmuir surfactant adsorption & dish room with 4 stations, 8 tasks $\cdot$ \real{datasheet three-compartment sink (Advance Tabco FS-3-1824-24RL)} $\cdot$ \real{T\&S B-0107 class 1.15 gpm spray valve, hose-reel valve, wrist-blade faucets, lever wastes} $\cdot$ 6 glasses, 4 to 6 plates, 2 bowls, fry pan, 15 cherry tomatoes, colander, 32 gal can $\cdot$ \real{listed products: Libbey rocks glass, Rubbermaid BRUTE can, Wear-Ever pan, Scotch-Brite sponge} & Mobile FR3 Duo: 2 FR3 arms, lift, swerve base $\cdot$ 2 robot setups $\cdot$ FR3 joint speed limits, collision reflexes $\cdot$ friction grasp up to 70 N per finger $\cdot$ valves and faucets turned by finger contact & \real{surfactant $\leq$ 10 $\mu$g per item, no 1 cm gauge $>$ 1 mg/m$^2$, no foam} $\cdot$ \real{visible soil $\leq$ 1 $\mu$g/cm$^2$ per 1 cm gauge} $\cdot$ $\geq$ 30 s immersed in sanitizer $\cdot$ free water $\leq$ 2--100 mL per item $\cdot$ placed in rack or on shelf, mouth-down or upright & 16 physics checks, e.g., Jeffreys film drainage, jet impact force vs $\rho$Qv, splatter vs Bhunia--Lienhard $\cdot$ 76 adversarial evaluator cases \\
\midrule
\multicolumn{6}{@{}l}{\textit{Liquid: wine pouring}} \\*
SoftGym\newline\citep{lin2020softgym} & position-based fluid & 1 liquid $\cdot$ box cups & none, cup moved directly & fraction poured & -- \\*
ManiSkill2\newline\citep{gu2023maniskill2} & MLS-MPM & 1 liquid $\cdot$ bottle $\cdot$ beaker & 1 Panda $\cdot$ end-effector deltas & level within 4 mm $\cdot$ under 100 particles spilt & -- \\*
\textbf{Ours} & DFSPH, 2 mm particles, 93,750 per bottle $\cdot$ cohesion $\cdot$ Young's-law wetting $\cdot$ lip separation $\cdot$ contact-line pinning $\cdot$ \real{headspace air and glugging} $\cdot$ added-mass coupling $\cdot$ absorbing optics & \real{4 liquids: red, white, and ros\'e wine, water} $\cdot$ \real{datasheet 75 cl bottle, Bordeaux, ISO tasting and lined glasses, carafes} & 2 FR3 arms $\cdot$ 4 robot setups $\cdot$ \real{Franka velocity, acceleration, and jerk limits} $\cdot$ friction grasp up to 70 N & volume in mL, e.g., 135--165 mL $\cdot$ spill and drips in mL $\cdot$ glass upright $\cdot$ bottle returned & 12 physics checks, e.g., hydrostatic load, sloshing frequency, tipping force $\cdot$ 12 adversarial evaluator cases \\
\midrule
\multicolumn{6}{@{}l}{\textit{Mechanism: key-and-padlock unlocking}} \\*
ManiSkill-ViTac\newline\citep{li2024maniskillvitac} & IPC contact with CCD $\cdot$ FEM sensor gels, Neo-Hookean $\cdot$ simulated GelSight Mini marker flow & 4 key--lock shape pairs $\cdot$ key teeth and lock pins & scene: 2 sensor gels and objects only $\cdot$ x, y, z displacement actions & key teeth to pin xyz error under threshold $\cdot$ gel surface points near initial $\cdot$ step limit & -- \\*
Meta-World\newline\citep{yu2020metaworld} & MuJoCo & door with rotating lock & 1 Sawyer arm $\cdot$ end-effector deltas and gripper torque & lock within 2 cm of goal & -- \\*
\textbf{Ours} & Newton 1.5 with MuJoCo Warp solver $\cdot$ native nonconvex triangle-mesh contact $\cdot$ 1,920 Hz, 64 solver iterations $\cdot$ 5-pin tumbler stacks: 5 key pins, 5 spring-loaded drivers $\cdot$ contact-only key discrimination, no key-code rule $\cdot$ round pin bores, 0.10 mm bore and 0.05 mm plug clearance $\cdot$ driver springs 40 N/m, about 0.3 N preload $\cdot$ lubricated brass friction 0.10 & 2 keys: correct 31542 and wrong 12345 $\cdot$ KW1-style 5-pin lock, 12.7 mm plug, 28 mm long $\cdot$ \real{KW1 bitting: 3.81 mm cut spacing, 0.5842 mm depth step} $\cdot$ \real{LAB/Kwikset pin sizes: 2.921 mm diameter, 4.3688--7.8740 mm key pins, 4.572 mm drivers} $\cdot$ \real{brass parts, 8,500 kg/m$^3$} & \real{1 Panda (franka\_description 0.7.1)} $\cdot$ \real{manufacturer joint limits, arm at 20\% rated speed} $\cdot$ \real{80 mm gripper stroke} $\cdot$ 45 N per-finger force cap $\cdot$ friction grasp, no key weld & latch released by plug turn past 30$^\circ$ with near key seated $\cdot$ key seated within 2 mm of keyway axis $\cdot$ shackle short leg over 6.2 mm clear of its bore (about 18$^\circ$ swing) $\cdot$ other key on its stand: within 8, 3, 3 mm and 20$^\circ$ $\cdot$ other key not gripped by both fingers & 13 physics checks, e.g., wrong-key rejection (3 cases), locked shackle under 5 N pull, 90$^\circ$ shackle swings both ways \\
\midrule
\multicolumn{6}{@{}l}{\textit{Metal: robotic MIG fillet welding}} \\*
Arc-Welding RL\newline\citep{xu2021deep} & 2D point kinematics, double integrator $\cdot$ welding as constant-speed traversal of an arc & 4 straight welding arcs $\cdot$ round obstacles & 3 round point robots $\cdot$ 2D control inputs & all arcs welded $\cdot$ no collision & -- \\*
Welding DDPG+IK\newline\citep{zhong2022collision} & CoppeliaSim collision detection $\cdot$ manipulator kinematics, Jacobian IK & obstacles in a semi-closed, narrow workspace & 1 6-DOF welding manipulator $\cdot$ joint-velocity actions & torch reaches target position $\cdot$ collision-free path & -- \\*
\textbf{Ours} & GPU enthalpy solver, 7.18 M cells, 0.25 mm in the joint band $\cdot$ \real{A36 melting and solidification, EN 1993-1-2 c\_p(T) and k(T)} $\cdot$ convection, radiation and table contact losses $\cdot$ \real{constant-voltage arc, melting law fit to Lincoln procedure charts} $\cdot$ \real{short-circuit, globular and spray transfer from the operating point} $\cdot$ \real{burnback, stubbing, 400 A source ceiling} $\cdot$ grounded welding circuit, touch sensing, stray arcs $\cdot$ \real{Young--Laplace pool surface per 0.5 mm slice: surface tension, gravity, arc pressure} & A36 base plate 200 $\times$ 150 $\times$ 10 mm and bracket 150 $\times$ 75 $\times$ 8 mm $\cdot$ 0.035 in ER70S-6 wire, 90/10 Ar/CO$_2$ gas $\cdot$ \real{datasheet equipment: Miller Auto-Continuum 350 source, wire feeder, gas cylinder} $\cdot$ \real{Siegmund-class fixture table, DESTACO toggle clamps, magnetic ground clamp} $\cdot$ \real{2.76 kg cobot MIG gun, 1.6 kg light robot torch} & UR10e with MIG gun, FR3 with Franka Hand $\cdot$ 6 robot setups incl. dual FR3, dual Panda, Unitree G1 $\cdot$ \real{UR10e 150 N force stop, libfranka 10/20 N m reflexes} $\cdot$ joint speed and acceleration limits $\cdot$ friction grasp, force mode up to 70 N & AWS D1.1 fillet size $\geq$ 6.35 mm on $\geq$ 90 \% of length $\cdot$ continuous fused weld end to end $\cdot$ throat, convexity, undercut and overlap limits $\cdot$ macro sections fused at root and both members $\cdot$ porosity $\leq$ 1.5 \% & 19 physics checks, e.g., Stefan solidification front, Rosenthal moving-source temperatures, arc current vs procedure charts \\
\midrule
\multicolumn{6}{@{}l}{\textit{Optics: barcode scanning}} \\*
RoboTwin 2.0\newline\citep{chen2025robotwin} & SAPIEN rigid-body simulation, 1/250 s step $\cdot$ convex collision shapes & 1 scanner, 5 models $\cdot$ 1 tea box, 6 models $\cdot$ clutter distractors in the randomized setting & 5 dual-arm embodiments: Aloha-AgileX, ARX-X5, Franka-Panda, Piper, UR5-Wsg $\cdot$ benchmark on Aloha-AgileX $\cdot$ one arm holds the scanner, the other the object & object within 25 mm of the scanner axis $\cdot$ object 0--70 mm in front of the scanner $\cdot$ both grippers closed & -- \\*
Sari Sandbox\newline\citep{gajo2025sari} & Unity with NVIDIA PhysX $\cdot$ rigid bodies with box colliders $\cdot$ ray cast to a barcode plane, orientation check & 250 grocery products in 11 categories $\cdot$ 3 store layouts $\cdot$ self-checkout counter with fixed scanner and touchscreen & avatar with camera and two hands $\cdot$ grip and poke toggles & manual visual inspection $\cdot$ task done within 45 min & -- \\*
\textbf{Ours} & MuJoCo via Newton 1.5 at 1 kHz, elliptic cones, converged no-slip pass $\cdot$ Coulomb and torsional friction per material $\cdot$ spring-return trigger and microswitch, 1.0 N + 0.45 N/mm $\cdot$ \real{Cherry MX Red force-travel on 17 keys} $\cdot$ path-traced imager frames at 30 Hz, 64--256 samples per pixel $\cdot$ \real{thin-lens defocus and diffraction, lens fitted to DS2208 decode ranges} $\cdot$ \real{645 nm LED illumination and 624 nm aim line} $\cdot$ \real{640 $\times$ 480 mono sensor, 32.8$^\circ$ $\times$ 24.8$^\circ$ field} & \real{DS2278-class scanner, 214 g, 175 $\times$ 66 $\times$ 90 mm datasheet envelope} $\cdot$ \real{CR2278 presentation cradle, 151 g} $\cdot$ 13 SKUs: cans, cereal, PET water bottle, glass jar, lip balm, batteries, packs, books $\cdot$ 12.5 kg case of water $\cdot$ \real{Uline H-7115-class boltless shelving with shelf-edge location labels} & 2 FR3 arms $\cdot$ 4 robot setups: 2 FR3, 2 Panda, 1 FR3, Unitree G1 with Dex3 hands $\cdot$ \real{Franka Hand force grasp up to 70 N (FR3), 40 N (Panda)} $\cdot$ friction grasp, nothing welded $\cdot$ coupled Franka Hand fingers & reads decoded from rendered frames and booked by the host app $\cdot$ every read traced to its printed label $\cdot$ no misreads or ghost reads $\cdot$ count per GTIN equals truth $\cdot$ each item read exactly once & 26 physics checks, e.g., trigger force-travel, decode range vs datasheet, drop vs analytic free fall $\cdot$ 12 adversarial evaluator cases \\
\midrule
\multicolumn{6}{@{}l}{\textit{Paint: walls, craft and spray}} \\*
ManiSkill3\newline\citep{tao2025maniskill3} & ink as pre-built thin cylinders placed under the tool $\cdot$ up to 1,000 ink dots per episode $\cdot$ GPU-parallel rigid-body simulation & white canvas on a table $\cdot$ goal triangle or SVG outline $\cdot$ 1 ink colour & 1 Panda with a stick tool $\cdot$ tool fixed to the arm & drawn dots within a distance of goal points $\cdot$ every goal point reached & -- \\*
FRIDA\newline\citep{schaldenbrand2023frida} & stroke appearance learned from real robot strokes $\cdot$ 3 stroke-shape parameters: thickness, bend, length $\cdot$ differentiable stroke compositing onto a canvas photo & 11 x 14 in canvas board $\cdot$ acrylic paints, up to 12 colours $\cdot$ brush, water and rag & 1 Rethink Sawyer $\cdot$ brush rigidly attached, perpendicular to canvas & -- & stroke appearance vs.\ real strokes \\*
\textbf{Ours} & Bingham thin-film lubrication on 1, 2 and 4 mm cell charts $\cdot$ \real{thixotropic Cross rheology with yield stress} $\cdot$ levelling, sag, wet edge and lap marks $\cdot$ \real{drying by room climate, Krieger-Dougherty build} $\cdot$ \real{Washburn absorption into porous substrates} $\cdot$ Kubelka-Munk optics, hiding judged by $\Delta$E*ab $\cdot$ compliant roller nap, 72 x 46 cells, film split $\cdot$ \real{sash brush: 144 pseudo-rigid-body tufts, 6,642 filaments} & 4 paints: interior latex eggshell, fluid acrylic, aerosol enamel, clear lacquer $\cdot$ Sherwin-Williams colours on tint bases, e.g., SW 6244 Naval over P-Shade primer $\cdot$ 5 applicators: 9 in roller, 2.5 in angled sash brush, 2 in dog-leg long-reach brush, \#8 round brush, 12 oz aerosol $\cdot$ \real{Wooster R404 tray, 1 qt cut pail, gallon can, six-well palette} $\cdot$ US bedroom 3.66 x 3.35 m: drywall, window and casing, baseboard, outlet, 16-section cast-iron radiator & Franka Mobile FR3 Duo: holonomic base, Spine lift, 2 FR3 arms, Franka Hands $\cdot$ 3 robot setups: mobile FR3 duo, mobile FR3 single, G1 with Dex3 hands for craft $\cdot$ friction pinch, 35 N per finger, 60 N for the roller $\cdot$ aerosol button opened by fingertip contact, about 23.5 N $\cdot$ protective-stop reflexes & hide: $\Delta$E*ab $\leq$ 1 to the paint's colour on 98 \% of a strip $\cdot$ no runs: $\leq$ 1 mm sag travel after settling $\cdot$ lap marks $\leq$ 1 cm2 $\cdot$ paint on ceiling, trim or radiator $\leq$ 0.5 cm2 $\cdot$ coat 25--50 $\mu$m on 90--95 \% & 19 physics checks, e.g., levelling vs thin-film theory, sag vs anti-sag notches, ASTM D5068 brush-out $\cdot$ 19 adversarial evaluator cases \\
\midrule
\multicolumn{6}{@{}l}{\textit{Paper: gift bags, wrap and cartons}} \\*
Deformable\-Ravens\newline\citep{seita2021learning} & PyBullet FEM mass-spring soft body $\cdot$ vertex self-collision $\cdot$ velocity-level contact and friction & 1 drawstring bag from a sphere mesh $\cdot$ 1 or 2 items $\cdot$ target zone & 1 UR5 with suction cup $\cdot$ grasp locks nearest vertex with fixed constraint & bead-ring convex hull area above threshold $\cdot$ items and bag opening in target zone & -- \\*
ThinShellLab\newline\citep{wang2024thinshell} & implicit Euler, Kirchhoff-Love shell stretching and bending $\cdot$ crease plasticity by yield angle $\cdot$ penalty contact with IPC smooth friction & 1 paper sheet on an arched table & 2 elastic manipulators, arm links omitted & middle crease angle sum $\geq$ success score & friction and bending fit to measured forces \\*
\textbf{Ours} & IPC barrier contact, implicit Newton solve, 4 ms step $\cdot$ \real{orthotropic St. Venant--Kirchhoff paper shells (MD/CD)} $\cdot$ \real{two-mechanism crease law with fold memory and ageing} $\cdot$ tearing by vertex splitting (VCCT), in-plane and out-of-plane $\cdot$ cohesive tape bonds with irreversible damage $\cdot$ compressible paper caliper contact $\cdot$ Coulomb friction with stick, per-material-pair coefficients $\cdot$ discrete elastic rod rope handles with knots & \real{laminated rope-handle Euro tote, 157 g/m$^2$ paper + 18 $\mu$m BOPP, on a 5-tote stack} $\cdot$ 3 bottom boards, boxed candle (420 g), rubber mat $\cdot$ \real{two-piece set-up gift box with telescoping lid and photo frame, 3 box sizes, 3 lid clearances} $\cdot$ \real{gift wrap in 2 stocks (81 and 53 g/m$^2$) with 3 strips of Scotch Magic 810 tape} $\cdot$ \real{RSC C-flute carton, 4 creased flaps, 48 mm 3M 371 seam tape, 8 tea cartons and a glass jar} & 2 FR3 arms with Franka Hands $\cdot$ 4 robot setups: 2 FR3, 2 Panda, 1 FR3, 1 Panda $\cdot$ Franka Hand force mode up to 35 N per finger $\cdot$ robot links as driven obstacles, coupling capped at 250 N $\cdot$ grasp by contact and friction, no attachment & bag mouth hull ratio $\geq$ 0.80 $\cdot$ bag standing within 10$^\circ$, bottom $\geq$ 85 \% flat $\cdot$ gussets open $\geq$ 120$^\circ$ $\cdot$ board in bag, candle upright on board $\cdot$ new tear $\leq$ 5 mm & 19 physics checks, e.g., incline friction stick and slide, lid air-film descent vs ODE, trouser tear force vs energy balance \\
\midrule
\multicolumn{6}{@{}l}{\textit{Paste: concrete, mortar and sealant finishing}} \\*
PlasticineLab\newline\citep{huang2021plasticinelab} & MLS-MPM $\cdot$ von Mises plasticity $\cdot$ grid contact with Coulomb friction & 1 plasticine box as dough $\cdot$ rigid rolling pin & none, 3-DoF kinematic capsule & normalized incremental IoU to target shape & -- \\*
DiffSkill\newline\citep{lin2022diffskill} & MLS-MPM (PlasticineLab) & 1 dough $\cdot$ spatula $\cdot$ rolling pin & none, velocity commands on tools & Sinkhorn/EMD to target below hand-picked threshold & -- \\*
\textbf{Ours} & Newton 1.5 MuJoCo rigid world, 0.5 ms ticks $\cdot$ 2.5-D paste height fields, 1--3 mm cells, 266,110 on the pad $\cdot$ \real{age-dependent yield stress anchored to slump and ASTM C403 set times} $\cdot$ thin-layer Bingham flow with yield gate $\cdot$ bleeding, trapped water and blisters $\cdot$ \real{ACI 308 evaporation by air and temperature} $\cdot$ Prandtl--Skempton bearing support plates $\cdot$ cut, roll, heave and squeeze-out & 5 pastes: concrete, repair mortar, Type N mortar, sealant, joint compound $\cdot$ \real{11 hand tools at maker sizes: floats, trowel, darby, edger, jointer, knives, penetrometer} $\cdot$ 5 stations: pad, floor patch, block joints, sealant joint, wallboard $\cdot$ flexible joint knife, fibreglass mesh patch $\cdot$ pump room with CMU walls, carts, hawk, mortar board, rag, tub & Mobile FR3 Duo: mobile base, spine lift, 2 FR3 arms $\cdot$ single-arm Mobile FR3 variant $\cdot$ base twist, joint targets, Cartesian deltas, gripper force $\cdot$ tools held by fingertip friction & pad within 3 mm of form plane $\cdot$ 1.2 m straightedge gaps $\leq$ 3 mm $\cdot$ 3/8 in edge radius within 1.5 mm on 90 \% of 2132 lines $\cdot$ 95 \% smooth-hard: densification $\geq$ 0.8, texture $\leq$ 0.05 mm $\cdot$ aggregate exposure $<$ 2 \% on 98 \% & 18 physics checks, e.g., float trench depth vs pitch, 8 cases, sealant tooled before vs after skin, 20 N vs 2 N mortar consolidation $\cdot$ 52 adversarial evaluator cases \\
\midrule
\multicolumn{6}{@{}l}{\textit{Rigid furniture: classroom rearrangement}} \\*
ManiSkill\newline\citep{mu2021maniskill} & SAPIEN with PhysX $\cdot$ articulated swivel chairs, up to 20 joints $\cdot$ wheel--ground contact & 36 swivel chairs (26 train, 10 test) from PartNet-Mobility $\cdot$ 1 chair per episode & mobile base with 2 Panda arms, 22 joints $\cdot$ PID joint or end-effector control $\cdot$ 3 robot-mounted cameras, 400 $\times$ 160 & chair within 15 cm of target $\cdot$ chair static for a period $\cdot$ chair not fallen over & -- \\*
iGibson 0.5\newline\citep{xia2020interactive} & common density and material friction for all objects $\cdot$ articulated doors as URDF joints $\cdot$ robot pushes interactable furniture & 106 scenes, 1,984 interactable CAD models: chairs, desks, doors, sofas, tables $\cdot$ 10 added scanned objects (baskets, shoes, pots, toys) at real-world weights & TurtleBot v2 and Fetch for baselines $\cdot$ 10 robot models available $\cdot$ joint position and velocity control & robot within its body width of the goal $\cdot$ 1,000 time steps $\cdot$ Interactive Navigation Score: path and effort efficiency & -- \\*
\textbf{Ours} & MuJoCo rigid multibody via Newton 1.5, 0.5 ms step $\cdot$ 56 free rigid furniture bodies with part-level mass, centre of mass and inertia $\cdot$ \real{pair-specific glide--floor friction for 3 glide types} $\cdot$ Stribeck static-to-sliding transition $\cdot$ tip vs slide set by push height and glide type $\cdot$ textbooks sliding and spilling from book boxes $\cdot$ swerve-drive base with wheel traction and drive-torque limits $\cdot$ FR3 joint-torque and hand-force collision reflexes & 71 m$^2$ U.S. secondary classroom (Room 114) $\cdot$ \real{24 Virco 785 desks, 28 Virco 9018 chairs} $\cdot$ tutoring table, teacher's desk, book tote, wastebasket $\cdot$ \real{listed product weights: desk 10.9 kg, chair 5.26 kg, table 32.3 kg, teacher's desk 62.6 kg} $\cdot$ \real{3 glide types: nylon, felt, steel} & \real{Franka Mobile FR3 Duo, 216.8 kg, from Franka's own description} $\cdot$ 2 FR3 arms with Franka Hands $\cdot$ holonomic TMR base with 2 ArgoDrive swerve units $\cdot$ \real{Franka Spine lift, 827 mm stroke} $\cdot$ \real{ArgoDrive 16 N$\cdot$m drive and 28.4 N$\cdot$m braking torque} & 10 tasks $\cdot$ every glide within 30 mm of its floor dot, yaw within 3$^\circ$ $\cdot$ chair tucked at least 0.30 m under the desk top $\cdot$ pod front edges -2 to +10 mm apart $\cdot$ exam chairs at least 1.25 m apart & 11 physics checks, e.g., glide breakaway and sliding friction, tip vs slide statics, 57 cases, book spill onset vs friction angle $\cdot$ 12 adversarial evaluator cases \\
\midrule
\multicolumn{6}{@{}l}{\textit{Snap-fit: pen capping}} \\*
Bi-DexHands\newline\citep{chen2022bidexhands} & Isaac Gym $\cdot$ 1 kHz low-level control & 1 pen with cap $\cdot$ SAPIEN dataset objects $\cdot$ cap on prismatic joint & 2 Shadow Hands, 24 DoF $\cdot$ movable hand bases & cap and body handles over 0.3 m apart & -- \\*
ALOHA Sim\newline\citep{zhao2023aloha} & MuJoCo & red peg, blue socket $\cdot$ about 5 mm clearance & 2 arms with grippers $\cdot$ ALOHA leader-arm teleoperation & peg touches pins inside socket & -- \\*
\textbf{Ours} & Newton 1.5 with MuJoCo Warp solver $\cdot$ native nonconvex contact: hollow SDF cap, triangle-mesh pen $\cdot$ 960 Hz, 30 solver iterations $\cdot$ 4 compliant snap nubs, 30 kN/m and 600 Ns/m each $\cdot$ 0.15 mm radial interference $\cdot$ 22$^\circ$ lead and 50$^\circ$ return ramps $\cdot$ contact-driven snap, shoulder stop and pull-off, no latch or weld $\cdot$ plastic friction 0.25 & 140 mm, 5.8 g pen with 5.25 mm bead $\cdot$ 58 mm, 5.0 g hollow cap, 5.40 mm bore, flared mouth $\cdot$ 0.22 kg movable notebook $\cdot$ workbench, robot pedestals, cabinetry $\cdot$ \real{D435-size camera on floor stand} & \real{2 Panda arms (franka\_description 0.7.1)} $\cdot$ \real{manufacturer joint and velocity limits} $\cdot$ 8 N per-finger effort limit $\cdot$ finger pad friction 0.80 $\cdot$ friction grasp, no weld, pen can pivot & writing tip inside cap bore: 0--56 mm deep, under 5.4 mm radial $\cdot$ all 4 nubs past bead crest edge (38.150 mm), not resting on corner $\cdot$ seated for the trailing 0.5 s $\cdot$ pen and cap each within 10 mm of final position $\cdot$ continuous progress score: depth and nubs snapped & 9 physics checks, e.g., snap insertion and pull-off force, notebook push vs Coulomb estimate, free-spin update vs analytic \\
\midrule
\multicolumn{6}{@{}l}{\textit{Soft tissue: fish filleting and meat trimming}} \\*
DiSECt\newline\citep{heiden2021disect} & FEM, Neo-Hookean tetrahedra $\cdot$ SDF knife contact $\cdot$ virtual-node cutting with weakening damage springs & 3 foodstuffs: apple, potato, cucumber $\cdot$ laser-scanned real meshes $\cdot$ knife, cutting board & none, knife on a prescribed velocity trajectory & cost: mean knife force and inverse velocity & knife forces vs.\ real cuts \\*
RoboNinja\newline\citep{xu2023roboninja} & MLS-MPM elastoplastic soft material $\cdot$ von Mises yield $\cdot$ SDF rigid core, knife and board & soft material around a rigid core $\cdot$ 300 training and 100 test objects $\cdot$ spline cores, 5 out-of-distribution core shapes & knife moved directly: $\delta$x, $\delta$y, $\delta$$\theta$ & completion: knife reaches the board $\cdot$ failure at $>$ 10 collisions or step energy $>$ 3.0 $\cdot$ cut mass ratio & -- \\*
\textbf{Ours} & explicit GPU FEM, stable neo-Hookean tets, 73,273 per trout $\cdot$ 1e-5 s substeps, 100 per 1 ms robot tick $\cdot$ \real{Atkins knife-edge cutting law, anisotropic toughness} $\cdot$ \real{blade sharpness index, honed and worn} $\cdot$ crack insertion along the blade's swept surface $\cdot$ cohesive interfaces: septum, rib cage, chain seam, silver skin $\cdot$ rigid bones with cuttable backbone joints $\cdot$ \real{rod bones that bend and break, 179 per trout} & 3 workpieces: rainbow trout, salmon half side, pork tenderloin $\cdot$ \real{trout 289.5 g with rigid head, 72 ribs, 62 pin bones} $\cdot$ \real{salmon side 971 g with 29 pin bones, belly fat strip, melanin spot} $\cdot$ tenderloin 536 g with chain, flake fat, silver skin $\cdot$ \real{Victorinox fillet knife, 173 $\times$ 25 mm blade} & 2 FR3 arms with Franka Hands $\cdot$ 3 robot setups: dual FR3, dual Panda, single FR3 $\cdot$ knife held by friction in channel inserts, 35 N per finger $\cdot$ \real{Franka Hand force mode up to 70 N} $\cdot$ \real{libfranka reflexes 10/20 N$\cdot$m, 100 N$\cdot$m for cutting} & head off through a backbone joint $\cdot$ fillet in one piece, yield $\geq$ 0.33 $\times$ gutted mass $\cdot$ no rib piece over 2 mm $\cdot$ NOAA workmanship $\geq$ 85 $\cdot$ fillet skin-down in tray, settled & 14 physics checks, e.g., press cut vs Atkins R$\cdot$$\ell$, toughness anisotropy across vs along sheets, notched-bar tearing strength \\
\midrule
\multicolumn{6}{@{}l}{\textit{Tool contact: whiteboard writing and erasing}} \\*
ManiSkill3\newline\citep{tao2025maniskill3} & PhysX GPU simulation (SAPIEN) $\cdot$ ink as pre-built thin cylinders moved under the tool $\cdot$ dot placed when tool tip is near canvas & white canvas on table $\cdot$ goal SVG outline, e.g., letter A $\cdot$ 1 ink colour, 1,000-dot budget & 1 Panda with stick tool $\cdot$ brush attached to robot & every drawn dot near the goal path $\cdot$ every goal-path point covered $\cdot$ threshold 0.05 m in docs, 0.1 m in code & -- \\*
Humanoid\-Bench\newline\citep{sferrazza2024humanoidbench} & MuJoCo $\cdot$ hinged wipe on tool rod & window $\cdot$ 1 window wiping tool & Unitree H1 humanoid $\cdot$ 2 Shadow Hands, 61 actuators $\cdot$ 448-taxel tactile grid & reward: wipe contact at 5 sites $\cdot$ reward: 0.5 m/s vertical wipe speed $\cdot$ reward: tool near both hands & -- \\*
\textbf{Ours} & Newton 1.5 MuJoCo rigid solver, 0.5 ms step $\cdot$ P1 tetrahedral nib and felt elasticity $\cdot$ \real{nib and felt moduli from QB/T 4164 Shore A and McMaster F13 felt} $\cdot$ P2 tetrahedral cap shell, 3 retention ribs $\cdot$ unilateral elastic contact solve $\cdot$ implicit regularized Coulomb board friction $\cdot$ gap-dependent magnetic eraser attraction $\cdot$ load-dependent ink transfer, 6 pigments, finite reservoirs & \real{6 STAEDTLER 351 markers: 138 mm, 19.6 g} $\cdot$ 6 colours, separate caps $\cdot$ \real{STAEDTLER 652 magnetic wiper, 107 $\times$ 57 $\times$ 20 mm} $\cdot$ \real{MooreCo 2.44 $\times$ 1.22 m porcelain-on-steel board} $\cdot$ \real{furnished classroom, movable chairs and desks at product heights} & floating Unitree G1 EDU, 29 body joints, no floor weld $\cdot$ 6 robot setups incl. Unitree H1-2 and Mobile FR3 Duo $\cdot$ 2 Dex3 three-finger hands, 14 joints $\cdot$ 43 joint position targets $\cdot$ \real{URDF effort limits, torque-speed envelope, Unitree rotor inertia} $\cdot$ friction grasp, no hand attachment & text: per-colour F1 $\geq$ 0.85, 7 $\times$ 7 texel window $\cdot$ bridge: 90\% coverage of every element $\cdot$ no uncovered run over 10 mm $\cdot$ stray ink $\leq$ 1 cm$^2$ $\cdot$ placement within 30 mm, 5\%, 2$^\circ$ & 15 physics checks, e.g., 27-case load, speed and tilt ink sweep, ink timestep and resolution convergence, magnetic eraser slide under gravity \\
\midrule
\multicolumn{6}{@{}l}{\textit{Tool use: drilling and anchoring}} \\*
Meta-World\newline\citep{yu2020metaworld} & MuJoCo rigid contact $\cdot$ nail on a slide joint & hammer $\cdot$ nail in wood block & 1 Sawyer $\cdot$ end-effector deltas, gripper torque & nail driven $>$ 0.09 m & -- \\*
DiSECt\newline\citep{heiden2021disect} & FEM, Neo-Hookean tetrahedra $\cdot$ SDF knife contact $\cdot$ damage springs for crack formation & knife $\cdot$ apple, potato, cucumber $\cdot$ laser-scanned meshes & knife on a prescribed velocity trajectory & -- & knife forces vs.\ real cuts \\*
\textbf{Ours} & bit feed from solved thrust, 1 ms tick $\cdot$ \real{Machinery's Handbook drilling law, steel and aluminium} $\cdot$ \real{measured wood cutting law, softwood and MDF} $\cdot$ \real{concrete removal by blow energy per volume: mortar, aggregate, rebar} $\cdot$ \real{push-gated SDS-plus blows with striker recoil} $\cdot$ \real{motor torque--speed lines, stall, overload clutch} $\cdot$ drill walking, chisel-edge threshold, pilots $\cdot$ \real{exit burrs and tear-out} & 11 tasks at 2 stations: concrete wall bay, drill-press bench $\cdot$ \real{C25/30 wall, 200 mm, hidden \#4 rebar mats} $\cdot$ \real{datasheet tools: DeWalt DCD800 drill, DCH273 rotary hammer, Grizzly G7943 press, Hilti SIW 6AT-A22 impact wrench} $\cdot$ vise, automatic centre punch, ball-peen hammer, shop vacuum, hole brush $\cdot$ \real{Hilti KB-TZ2 wedge anchors, M8 and carriage bolts} & 2 FR3 arms on a mobile cart $\cdot$ 4 robot setups $\cdot$ \real{Franka velocity, acceleration, and jerk limits} $\cdot$ \real{87 and 12 N$\cdot$m joint torque caps} $\cdot$ FR3 collision reflex & hole position, e.g., within 0.5 mm of the mark $\cdot$ axis within 1--3$^\circ$ of the face normal $\cdot$ depth, e.g., 108--125 mm $\cdot$ hole diameter and roundness $\cdot$ rebar not cut ($\leq$ 0.5 mm damage) & 16 physics checks, e.g., metal thrust and torque vs published data, blow share vs push, clamp slip vs hand calculation \\
\midrule
\multicolumn{6}{@{}l}{\textit{Tool use: screw tightening}} \\*
Factory\newline\citep{narang2022factory} & SDF collisions $\cdot$ contact reduction $\cdot$ Gauss-Seidel solver in PhysX & M4--M20 nuts and bolts $\cdot$ 60 NIST Task Board 1 part models & 1 Franka $\cdot$ 7 controllers $\cdot$ parallel-jaw grasp of the nut & nut within 1 thread of the bolt base & fingertip forces vs.\ human data \\*
RLBench\newline\citep{james2020rlbench} & V-REP/PyRep simulation $\cdot$ nail turned on a joint & screwdriver $\cdot$ nail in a block & 1 Panda $\cdot$ joint, torque, or end-effector actions $\cdot$ grasp by attachment (default) & screwdriver grasped $\cdot$ nail joint turned 1.4 rad (about 90$^\circ$) & -- \\*
\textbf{Ours} & \real{ISO 16047 thread and under-head friction from solved loads} $\cdot$ \real{helical screw joints, 0.5 and 0.7 mm pitch} $\cdot$ self-locking threads, Coulomb stick and breakaway $\cdot$ one-way detent clutch: ramp, click, notch $\cdot$ clamp load through head-bearing stiffness $\cdot$ reflected axial screw inertia $\cdot$ \real{clamp friction for 7 material pairs} $\cdot$ cantilevered rail droop on a loose screw & \real{SICK W4F-size sensor on BEF-W4-A bracket, 30$^\circ$ arc slot} $\cdot$ 4 prestarted socket-head screws: 2 M3, 2 M4 $\cdot$ \real{20 mm aluminium frame: HFS5-2020 rail, post, 15 g corner bracket} $\cdot$ \real{2 preset torque drivers: Wera 7440-type 0.35 N$\cdot$m, HansTool 0176-type 0.60 N$\cdot$m} $\cdot$ rubber V-groove holders, sliding reference flag, service guard & 2 FR3 arms $\cdot$ 3 robot setups $\cdot$ Franka Hand, 70 N finger force $\cdot$ friction pinch grasp of the driver $\cdot$ joint, gripper, and Cartesian actions & driver clutch clicked with bit in socket $\cdot$ $\geq$ 80 \% of equation clamp load retained, e.g., 288 N for M3 $\cdot$ \real{M3 bit moment $\leq$ 0.4 N$\cdot$m (SICK limit)} $\cdot$ rail level and square within 3 mrad $\cdot$ sensor ray detects the flag & 10 physics checks, e.g., click load vs ISO torque equation, clamped breakaway vs friction capacity, loose-rail droop vs hand estimate \\
\midrule
\multicolumn{6}{@{}l}{\textit{Vehicle: humanoid forklift operation}} \\*
Forklift Sim2Real\newline\citep{oishi2025visual} & Isaac Sim articulation-based drivetrain controllers $\cdot$ drive speed response tuned to a real 1/14-scale forklift & 1/14-scale counterbalance forklift, simplified model $\cdot$ 1 pallet with load on a stand $\cdot$ 1.8 m square walled space & none: the forklift is the agent $\cdot$ normalized throttle and steering actions $\cdot$ 2 side cameras, 352 $\times$ 288 RGB & forks reach the pallet position (reward term) $\cdot$ penalty if the pallet moves $\cdot$ real trials: lift, reverse and lower without dropping the load & -- \\*
DARPA VRC\newline\citep{aguero2015inside} & ODE maximal-coordinate rigid bodies $\cdot$ torque-controlled pin joints with position, velocity and torque limits $\cdot$ sphere, box and cylinder collision primitives & Polaris Ranger EV utility vehicle $\cdot$ roads, mud and debris & 1 Atlas humanoid, 28 DoF $\cdot$ Sandia hands: 4 fingers, 12 DoF $\cdot$ vehicle worked through steering wheel, hand brake, forward-reverse switch and pedals & 4 ordered checkpoints: enter vehicle, drive 2 road halves, walk to finish $\cdot$ drive through gates $\cdot$ run ends after 3 falls & -- \\*
\textbf{Ours} & MuJoCo rigid multibody via Newton 1.5, 0.5 ms step, 21-body truck $\cdot$ \real{super-elastic tyres: 3 ribs per tyre, p\^{}1.5 contact law, Maxwell rubber damping} $\cdot$ \real{SUPREM brush tyre model with relaxation and Coulomb stick-slip} $\cdot$ \real{torque-converter powershift drivetrain: engine torque curve, clutch packs, open differential} $\cdot$ \real{orbitrol hydrostatic steering with 12 MPa relief, leakage and Ackermann linkage} $\cdot$ \real{staged triplex mast hydraulics: compressible oil columns, 18.5 MPa relief, lowering regulator, cylinder leakage} $\cdot$ oscillating steer axle and three-point stability $\cdot$ dry-friction stick/slip service and parking brakes & \real{2.5 t LPG counterbalance forklift: Isaac Sim Forklift C re-rigged with Hyster H2.5A masses and drivetrain} $\cdot$ 11 operator controls: 3 pedals, 3 mast levers, direction stalk, steering wheel and spinner knob, key, horn, parking brake $\cdot$ 3 presets: racked warehouse, 4-door shipping dock, yard with mobile ramp $\cdot$ \real{GMA 48 $\times$ 40 in pallets with wrapped loads of 600--1,200 kg} $\cdot$ steel coils, loose carton stack (5 layers of 6), OCC bales of 450--590 kg & 1 Unitree H1-2 with Inspire hands, seated $\cdot$ held by contact and lap belt, no weld $\cdot$ 39 motors on Unitree LowCmd channels (q, dq, kp, kd, tau) $\cdot$ URDF torque--speed envelope per joint $\cdot$ controls moved only by contact, nothing attached & 14 tasks in 3 presets $\cdot$ placement vs zone geometry: bay lines, rack beams, conveyor rails, trailer nose $\cdot$ rack overhang 51--101 mm, pallet level within 2$^\circ$ $\cdot$ no strike above 2,000 N or 50 N$\cdot$s $\cdot$ no wheel lift over 0.3 s, roll under 6$^\circ$, no overturn & 15 physics checks, e.g., tilt-table tipping angle, 0--15 m time and top speed vs datasheet, coast-down on grades vs theory \\
\bottomrule
\end{longtable}
}
\section{Additional Details on Learning in Simulation}
\label{app:learning-results}

\paragraph{An example of coaching.} In towel folding, GPT-6 Astra attempted the task as a coached student with restricted access. After more than four and a half hours and five hints, it still could not hold the towel with both grippers at once, and asked its coach for a single qualitative change. The coach replied:
\begin{quote}
\small ``Try a staged fold using one gripper: place one end of the long edge onto its opposite counterpart, release onto the table, then regrasp and bring the remaining section over. Keeping most fabric supported reduces the load on each grasp and avoids needing two sustained holds.''
\end{quote}
The student then folded the towel with its left gripper alone, while its right arm stayed idle. Its score rose with each following action, and it succeeded about 40 minutes after the hint.

\section{Extended Related Work}
\label{app:related}

This appendix expands on the related work in \S\ref{sec:related}.

\paragraph{Simulation for robot learning.} Simulation has long been a central tool for robot learning, and physics engines such as MuJoCo~\citep{todorov2012mujoco}, Isaac Gym~\citep{makoviychuk2021isaac}, ManiSkill3~\citep{tao2025maniskill3}, and Genesis~\citep{genesis2024} make it fast, with GPU-parallel engines running thousands of environments at once. Benchmarks built on these engines cover tabletop manipulation of rigid and articulated objects~\citep{yu2020metaworld,james2020rlbench,liu2023libero}, household environments~\citep{li2021igibson,nasiriany2024robocasa,nasiriany2026robocasa365}, and specific workplaces such as laboratories~\citep{li2025labutopia,lan2026autobio} and retail stores~\citep{gajo2025sari}. A separate line of benchmarks targets deformable materials, such as liquids~\citep{xian2023fluidlab,li2024chemistry3d,gong2023arnold}, cloth and garments~\citep{seita2021learning,lu2024garmentlab,wang2025dexgarmentlab,hu2025rgbench}, plasticine~\citep{huang2021plasticinelab,lin2022diffskill}, or several of these~\citep{lin2020softgym,gu2023maniskill2,chen2023daxbench,li2026lehome}. BEHAVIOR-1K~\citep{li2023behavior1k} comes closest to broad coverage, simulating both liquids and cloth and grounding its 1,000 household activities in time-use surveys. However, these benchmarks are built mostly by hand, so each covers only the materials and activities that its authors chose and could implement, and adding a new kind of physics requires substantial expert effort. Moreover, none of the benchmarks in Table~\ref{tab:benchmarks} selects its tasks or simulations by the economic value of physical work, and only RGBench~\citep{hu2025rgbench}, which targets garments, checks its simulated physics against real measurements. In contrast, Gym-Anything for Robotics uses coding agents to build each simulation, including physics that existing engines lack, selects simulations based on occupation and time-use data, and checks the physics of each simulation against physical theory and, where available, real measurements.

\paragraph{Reliable physics in simulation.} Whether simulated physics can be trusted is a long-standing question in robotics, since policies trained in simulation degrade when its physics differs from the real world. A large body of work narrows this sim-to-real gap by identifying physical parameters from real data~\citep{tan2018simtoreal}, randomizing them during training~\citep{tobin2017domain,peng2018simtoreal}, or reconstructing real scenes in simulation~\citep{torne2024rialto}, while specialized solvers model materials that general-purpose engines handle poorly, such as liquids~\citep{xian2023fluidlab}, thin shells~\citep{wang2024thinshell}, and cutting~\citep{heiden2021disect}. These methods can produce faithful physics for a specific task, but each requires experts to model and tune the physics by hand. Recent benchmarks also measure this gap directly, by comparing simulated cloth with real cloth manipulated by robots~\citep{blancomulero2024benchmarking,hu2025rgbench}, but they cover a single material. A more recent line of work uses LLMs to generate simulations, writing task code~\citep{wang2024gensim,hua2024gensim2}, demonstration programs~\citep{chen2025robotwin}, scenes~\citep{yang2024holodeck,wang2024robogen}, or reward functions~\citep{ma2024eureka}. However, these methods generate tasks and scenes on top of the fixed physics of an existing engine, which largely limits them to rigid and articulated objects, and they check their output through policy success, visual inspection, or human review rather than by measuring the simulated physics. Our work builds on the creation-audit loop of Gym-Anything~\citep{aggarwal2026gymanything}, which verifies environments built from real software, and extends it to physics, where coding agents implement the physics itself and audit agents check it through physical measurements, assisted demonstrations, and adversarial robot actions.

\paragraph{Frontier models learning in simulation.} Frontier models increasingly act in simulated and physical worlds, either by writing programs that call perception and control primitives~\citep{liang2023code,huang2023voxposer} or by controlling robots directly~\citep{geminirobotics2025}, and several benchmarks evaluate multimodal models as embodied agents~\citep{yang2025embodiedbench,zhang2024vlabench}. Such models can also improve without weight updates, by learning from repeated attempts in context~\citep{brown2020language,shinn2023reflexion} or by building a library of skills as code, as Voyager~\citep{wang2023voyager} does in Minecraft. Separately, robot learning has long used the privileged state of simulators, both by training a teacher with access to the true state and distilling it into a student that observes only sensors~\citep{chen2020learning,lee2020learning,pinto2018asymmetric}, and by generating demonstrations in simulation at scale to train visuomotor policies~\citep{mandlekar2023mimicgen,nasiriany2024robocasa}. However, these two lines of work rarely meet: privileged teachers pass their knowledge to students through weight updates, while frontier-model agents have been studied mainly in games and rigid-object scenes, without privileged state as a learning signal. We bring the two together on realistic simulations, showing that privileged state accelerates the in-context learning of frontier models, that a privileged frontier model can teach a student through qualitative hints alone, that a model can learn from its own privileged experience, and that privileged code policies can train VLAs that observe only cameras and proprioception.

\paragraph{Economic grounding of benchmarks.} A growing body of work connects AI capabilities to the economy through occupation data, most prominently by estimating which occupations and tasks in O*NET~\citep{peterson2001onet} are exposed to language models~\citep{eloundou2024gpts}. These studies measure exposure rather than guide the design of benchmarks, but recent benchmarks have begun to use economic value directly: GDPval~\citep{gdpval2025} evaluates models on tasks from 44 occupations across the nine sectors that contribute most to U.S.\ GDP, and Gym-Anything~\citep{aggarwal2026gymanything} selects software for computer-use environments by its attributed GDP. In robotics, BEHAVIOR-1K~\citep{li2023behavior1k} grounds its household activities in time-use surveys, but to our knowledge no prior robotics simulation benchmark draws on occupation data. Gym-Anything for Robotics extends GDP-grounded selection to physical work, combining the tasks and wages of each occupation with the time that people spend on everyday activities~\citep{bls2025atus}, and selects simulations that supply the physical capabilities that the most valuable actions require.

\end{document}